\documentclass[times, review, 10pt]{elsarticle}
\usepackage{amsmath,amsfonts}
\usepackage{algorithmic}
\usepackage{algorithm}
\usepackage{array}
\usepackage[caption=false,font=normalsize,labelfont=sf,textfont=sf]{subfig}
\usepackage{textcomp}
\usepackage{stfloats}
\usepackage{url}
\usepackage{verbatim}
\usepackage{graphicx}
\usepackage{tabularx}                      

\usepackage{amsmath,amssymb,amsfonts}
\usepackage{algorithm}
\usepackage{algorithmic}
\usepackage{graphicx}
\usepackage{textcomp}
\usepackage{xcolor}
\def\BibTeX{{\rm B\kern-.05em{\sc i\kern-.025em b}\kern-.08em
    T\kern-.1667em\lower.7ex\hbox{E}\kern-.125emX}}

\usepackage{times}
\usepackage{soul}
\usepackage{url}
\usepackage{multirow}
\usepackage{amsfonts}
\usepackage{balance}
\usepackage{graphicx}
\usepackage{lscape}
\usepackage{booktabs}
\usepackage{makecell}
\usepackage{bbm}
\usepackage{amsthm}
\usepackage{multirow}

\usepackage{xcolor}
\usepackage{dirtytalk}
\usepackage[T1]{fontenc}
\usepackage{array} 
 
\usepackage{amssymb}
\usepackage{amsmath}

\journal{Pattern Recognition}

\begin{document}

\begin{frontmatter}



\title{A Multi-modal Fusion Network for Terrain Perception Based on Illumination Aware}




\author[1]{Rui Wang}
\ead{bhwangr@buaa.edu.cn}
\author[2]{Shichun Yang}
\ead{yangshichun@buaa.edu.cn}
\author[1]{Yuyi Chen}
\ead{yychen@buaa.edu.cn}
\author[1]{Zhuoyang Li}
\ead{18374167@buaa.edu.cn}
\author[1]{Zexiang Tong}
\ead{tzzxxx@buaa.edu.cn}
\author[1]{Jianyi Xu}
\ead{zy2457928@buaa.edu.cn}
\author[3]{Jiayi Lu}
\ead{lujiayi@buaa.edu.cn}
\author[1]{Xinjie Feng}
\ead{bhfengxinjie@buaa.edu.cn}
\author[3,4]{Yaoguang Cao\corref{cor1}}
\ead{caoyaoguang@buaa.edu.cn}

\cortext[cor1]{Corresponding author: Yaoguang Cao. Email: caoyaoguang@buaa.edu.cn}

\address[1]{Department of Transportation Science and Engineering, Beihang University, Beijing, China.}
\address[2]{Department of Transportation Science and Engineering, Beihang University, and Innovation Center of New Energy Vehicle Digital Supervision Technology and Application for State Market Regulation, Beijing, China.}
\address[3]{Hangzhou International Innovation Institute, Beihang University, Hangzhou, China.}
\address[4]{State Key Lab of Intelligent Transportation System, Beihang University, Beijing, China. }

\fntext[fn1]{This work was supported by the National Key R\&D Program of China, No: 2022YFB3206600.}

\begin{abstract}
Road terrains play a crucial role in ensuring the driving safety  of autonomous vehicles (AVs). However, existing sensors of AVs, including cameras and Lidars, are susceptible to variations in lighting and weather conditions, making it challenging to achieve real-time perception of road conditions. In this paper, we propose an illumination-aware multi-modal fusion network (IMF), which leverages both exteroceptive and proprioceptive perception and optimizes the fusion process based on illumination features. We introduce an illumination-perception sub-network to accurately estimate illumination features. Moreover, we design a multi-modal fusion network which is able to dynamically adjust 
weights of different modalities according to illumination features. We enhance the optimization process by pre-training of the illumination-perception sub-network and incorporating illumination loss as one of the training constraints. Extensive experiments demonstrate that the IMF shows a superior performance compared to state-of-the-art methods. The comparison results with single modality perception methods highlight the comprehensive advantages of multi-modal fusion in accurately perceiving road terrains under varying lighting conditions. Our dataset is available at: https://github.com/lindawang2016/IMF.
\end{abstract}

\begin{graphicalabstract}
\end{graphicalabstract}

\begin{highlights}
\item  We propose an illumination-aware multi-modal fusion network (IMF) that leverages both exteroceptive and proprioceptive data to enhance road terrains perception under varying light conditions.
\item Illumination features are incorporated into the fusion process, allowing dynamic adjustment of modality weights to improve perception under different conditions.
\item We construct two sets of multi-modal fusion percetion system and conduct extensive experiments, evaluating the effectiveness of the proposed algorithm.
\end{highlights}


\end{frontmatter}



\section{Introduction}

Autonomous driving technology has reached swift advancement, featuring the incorporation of various sensors, including cameras ~\cite{AVcamera}and LiDARs~\cite{AVLIDAR}, alongside of deep learning algorithms~\cite{AVDL1}. However, current autonomous driving research has shown limited focus on road surface conditions which have a significant impact on the driving safety of AVs. In fact, as pointed out by the World Road Association ~\cite{PIARC2024},  \say{\emph{Road infrastructure is strongly linked to fatal and serious injury causation in road collisions}}. Different types of road surfaces (e.g. wet, muddy, gravel or asphalt) can have a significant effect on the vehicle's driving stability~\cite{roadsurface}, braking distance and handling. For instance,  real-time perception of road surface types enables AVs to optimize the anti-lock braking system (ABS) parameters~\cite{roadABS}, such as the optimal slip rate, ultimately reducing braking distance and mitigating collision risksy. Therefore, it is essential for AVs to actively perceive road surface types to ensure driving safety.

The existing research on road condition recognition can be categorized into two types: exteroceptive perception and proprioceptive perception methods~\cite{exteroproprio}. Exteroceptive sensors, such as cameras~\cite{roadcamera1} and LiDARs~\cite{roadradar}, sense the terrain from a distance and enable the vehicle to classify its surroundings without directly interacting with it~\cite{exteroproprio}. However, these sensors are susceptible to weather and lighting conditions, complicating accurate perception across diverse environments~\cite{imageweather1}. Moreover, the substantial cost of LiDAR limits widespread deployment in vehicles~\cite{badlidar}. Proprioceptive methods, including accelerometers~\cite{roadacc1} and intelligent tires~\cite{meroadtire}, sense terrain properties through the interaction of the vehicle with its environment and their data can be used to train accurate terrain classifiers~\cite{roadtire2}. 

 Given the dynamic changes of autonomous driving scenarios, relying on a single modality to capture all road surface features proves challenging~\cite{road_single}. Recent multi-modal deep learning research has demonstrated the potential to learn complementary features~\cite{PR3}, prompting us to adopt a similar approach by fusing two modalities for robust road surface perception.

Currently, multi-modal fusion methods can be categorized as aggregation based, alignment based, and channel-exchange based approaches ~\cite{CEN}. Among these, channel-exchange based methods, which facilitate directional exchange of information across specific channels within each modality, have shown significant advantages  across multiple research domains, such as disease recognition~\cite{PRfusion}, remote sensing ~\cite{remotesensing} and semantic segmentation~\cite{PRfusion2}. Extensive research has demonstrated that these methods enhance fusion performances, outperforming aggregation-based~\cite{concat1} and alignment-based techniques~\cite{MMD}. We believe that channel-exchange fusion methods are well-suited for extracting complementary features from different modality data, thereby improving the accuracy of road surface condition perception. Motivated by the excellent performance of multi-modal fusion, ~\cite{shifusion} proposed the visual-tactile fusion method that integrates tactile information between vehicles and roads surface with images for road condition perception.

While research in multi-modal fusion has made significant progress in road condition perception, there are still some issues that have not been thoroughly investigated. One of the key issues is the impact of ambient lighting on camera-based perception~\cite{Lowlight}, which can significantly degrade performance under low-light or extreme lighting conditions. Existing multi-modal fusion approaches in AVs~\cite{fusionAV_review}, often treat all sensor modalities with fixed or implicitly learned fusion weights while do not fully discuss method's performance on different light conditions. For example, ~\cite{fusionAV4} focuses on the scene understanding performance of the algorithm but does not discuss the impact of different environmental lighting conditions on the results. However, studies have shown that the human brain dynamically reweights sensory inputs depending on environmental conditions~\cite{humanbrain}. Additionally, ~\cite{PIA,illuminationinvariant} has found that compared to implicit modeling of illumination, explicit modeling can better adapt to varying lighting conditions and reduce reliance on training data. Inspired by these findings, we argue that multi-modal fusion for AVs should also explicitly account for variations in lighting conditions.


To tackle these challenges about road condition perception, we propose a multi-modal fusion network based on illumination aware, which utilize proprioceptive sensors to compensate for the limitations of exteroceptive perception in low-light scenarios. Specially, we design an illumination perception sub-network that takes image data as the input and extract illumination features across different light conditions. Furthermore, we propose a multi-modal fusion network to optimize the integration of  exteroceptive and  proprioceptive data according to the illumination features. By employing a squeeze-excitation (SE) mechanism, the network dynamically allocates weights to different modalities according to the prevailing lighting conditions. We also adapt the training procedure of the road perception network. We also enhance the training process by pre-training the illumination-perception sub-network and incorporating illumination loss into the overall training objective.

In order to facilitate this work, we build up two types of multi-modal fusion perception system: one equipped with a camera and an accelerometer mounted on the vehicle suspension, and the other utilizing a camera and intelligent tires. Data from both exteroceptive and proprioceptive modality is collected under different lighting conditions and vehicle speeds. Extensive experiments have proved that our proposed method has superior performance than other baselines under various lighting and driving conditions. 

To sum up, our major contributions are three-fold:
\begin{itemize}
\item We propose a novel illumination-aware multi-modal fusion network that enables accurate perception of road terrains under varying lighting conditions.
\item We build up two types of multi-modal fusion perception systems and create two sets of multi-modal dataset that include exteroceptive and proprioceptive data collected under varying illumination levels and vehicle speeds.
\item Extensive experiments demonstrate that the superiority of our proposed algorithm over other state-of-the-art algorithms. Compared with single modality perception methods, the adopted visual-tactile fusion method can leverage the complementary information of two modalities under different lighting conditions. 
\end{itemize}

The remainder of this paper is organized as follows. The illumination-aware perception applications are also discussed in this section. In Section~\ref{Sec:Proposed}, we introduce our proposed network IMF in details, including the problem definition, the network architecture and the optimization process. In Section~\ref{Sec:results}, we discuss the perception results of our method in comparison to other baseline methods as well as in comparison to single-modality perceptual methods. Section~\ref{Sec:conclusion} briefly concludes some remarks and future works.

\section{Methdology}
\label{Sec:Proposed}
\subsection{Problem definition} In our problem, all the training data contains two modalities of data, exteroceptive and proprioceptive data. During the training process, the input data are integrated as multimodal pairs $\{\mathbf{x}_e^{i},\mathbf{x}_p^{i}\}$ with road type labels $\{\mathbf{y}_{r}^{i}\}$. Our goal is to find a multi-modal fusion network ${f}_{r}$ whose output $\{\mathbf{\hat{y}}_{r}^{i}\}$ is expected to fit $\{\mathbf{y}_{r}^{i}\}$ as close as possible. This can be achieved by minimizing the empirical loss as shown in Eq.~\ref{equation:claloss}:
\begin{equation}
\min _{{f}_{r}} \frac{1}{N} \sum_{i=1}^{N} \mathcal{L}_{r}\left(\hat{\mathbf{y}}_{r}^{i}={f}_{r}\left(\mathbf{x}_e^{i},\mathbf{x}_p^{i}\right), \mathbf{y}_{r}^{i}\right)
\label{equation:claloss}
\end{equation}

Considering the effect of ambient illumination on the fusion process, we also assign illumination condition labels $\{\mathbf{y}_{i}^{i}\}$ to the visual-tactile fusion data pairs. We first estimate the lighting conditions through the illumination perception sub-network  $f_{i}$ and compute the illumination features. This sub-network can be optimized by Eq.~\ref{equation:lightloss}:
\begin{equation}
\min _{f_{i}} \frac{1}{N} \sum_{i=1}^{N} \mathcal{L}_{i}\left(\hat{{y}}_{i}^{i}=f_{i}\left(\mathbf{x}_e^{i}\right), {y}_{i}^{i}\right)
\label{equation:lightloss}
\end{equation}

\subsection{Architecture}
\begin{figure*}[htp]
    \centering
    \includegraphics[width=12cm]{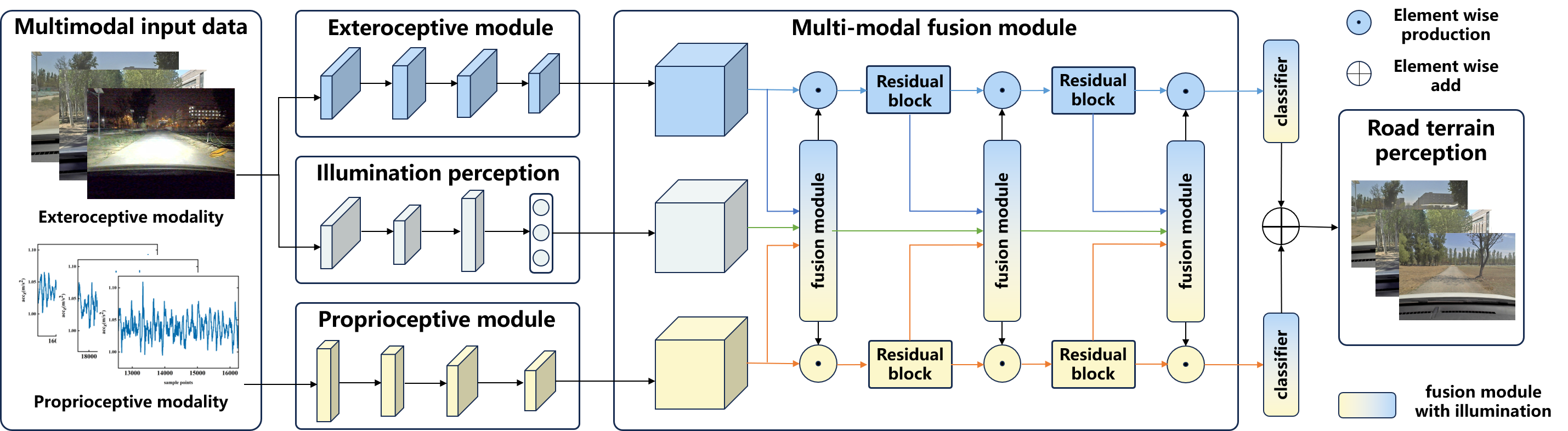}
    \caption{The proposed illumintion-aware multi-modal fusion network}
    \label{fig:network}
\end{figure*}

In order to achieve accurate road terrains perception under different light conditions, this paper optimizes the multi-modal fusion process through three steps: (a):  utilizing an illumination perception sub-network to obtain illumination features; (b): introducing illumination features into the multi-modal fusion module to adjust the attention weights of two modalities by the SE attention mechanism; and (c): enhancing the training process by pre-training the illumination perception sub-network and integrating illumination loss into the overall loss function. Apart from the mentioned modules, the proposed multi-modal fusion algorithm includes feature extractors for both modalities and a classifier for terrain types. The overall architecture is shown in Fig~\ref{fig:network}, with detailed discussions of each module provided in the following sections. 

\begin{figure}[htp]
    \centering
    \includegraphics[width=
    7cm]{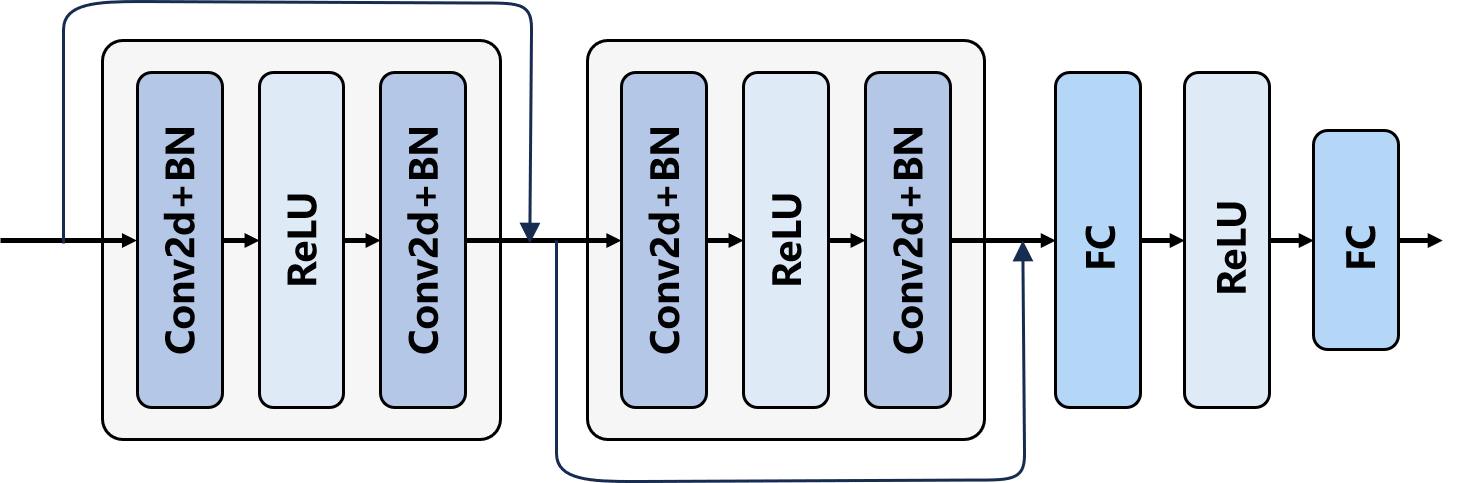}
    \caption{The illumination perception module}
    \label{fig:light}
\end{figure}

\textbf{Illumination perception module:} In order to adjust the multi-modal fusion process according to lighting conditions, we first design an illumination perception sub-network $f_{i}$ to extract the illumination features. This sub-network consists of a feature extractor and a classifier to estimate the lighting conditions. The feature extractor takes the image data as input and contains two layers of residual module, which have been proven to have strong feature extraction capabilities~\cite{resnet}. The sub-network outputs the estimated illumination features $\mathbf{F}_{i}$ with true values assigned as  1, 0.5, 0 for day, dusk, and night. The details of the illumination perception module is demonstrated as shown in  Fig.~\ref{fig:light}.

\textbf{Feature extractor:} We first designed feature extractors $f_{G}$, which contains both exteroceptive and proprioceptive module to perform preliminary feature extraction on multi-modal data pairs $\left\{\mathbf{x}_{e}^{i}, \mathbf{x}_{p}^{i}\right\}$, outputing initial feature representations $\mathbf{F}_{e}$ and $\mathbf{F}_{p}$. In this module, convolutional layers are used to extract features and downsize the data. The BatchNorm (BN) layers are  utilized to accelerate the training and convergence of the network, control the gradient explosion and prevent gradient vanishing~\cite{BN}. We also use the pooling layer, which is proved
to speed up the computation and prevent overfitting~\cite{poolinglayer}. The feature extractor is designed for exteroceptive and proprioceptive modalities respectively and details are demonstrated in Table ~\ref{table:feature extractor}.

\begin{table}
\scriptsize
\centering
\caption{Feature Extractors $f_{G}$ for multi-modal data}
\begin{tabular}{cccc}
\hline
\textbf{ Exteroceptive module}                                    & \textbf{Output}               & \textbf{ Proprioceptive module}                                    & \textbf{Output}               \\ \hline
conv2d(3,64,7,3,3) & (bs,64,86,86) & conv2d(3,64,7,3,3) & (bs,64,86,86) \\ 
BatchNorm2d(64)                                  & (bs,64,86,86) & BatchNorm2d(64)                                  & (bs,64,86,86) \\ 
ReLU()                                           & (bse,64,86,86) & ReLU()                                           & (bs,64,86,86) \\ 
MaxPool2d(3,3,1)     & (bs,64,29,29) & MaxPool2d(3,3,1)     & (bs,64,29,29) \\ \hline
\end{tabular}
\label{table:feature extractor}
\end{table}

\textbf{Multi-modal fusion module:} Considering the effect of lighting conditions on the multi-modal fusion, we take the illumination features into the fusion process explicitly. First, the exteroceptive and proprioceptive features $\mathbf{F}_{e}$ and $\mathbf{F}_{p}$ are fed into the residual layers for further feature extraction as shown in Eq.~\ref{equation:residual}.

\begin{equation}
\mathbf{F}_{e}^{' l}=R L_{e}\left(\mathbf{F}_{e}^{l}\right) \quad \mathbf{F}_{p}^{' l}=R L_{p}\left(\mathbf{F}_{p}^{l}\right)
\label{equation:residual}
\end{equation}
where $l$ represents the layer number of the multi-modal fusion module and $RL_{e}$ and $RL_{p}$ are residual layers for exteroceptive and proprioceptive modality, respectively.

\begin{figure}[htp]
    \centering
    \includegraphics[width=5cm]{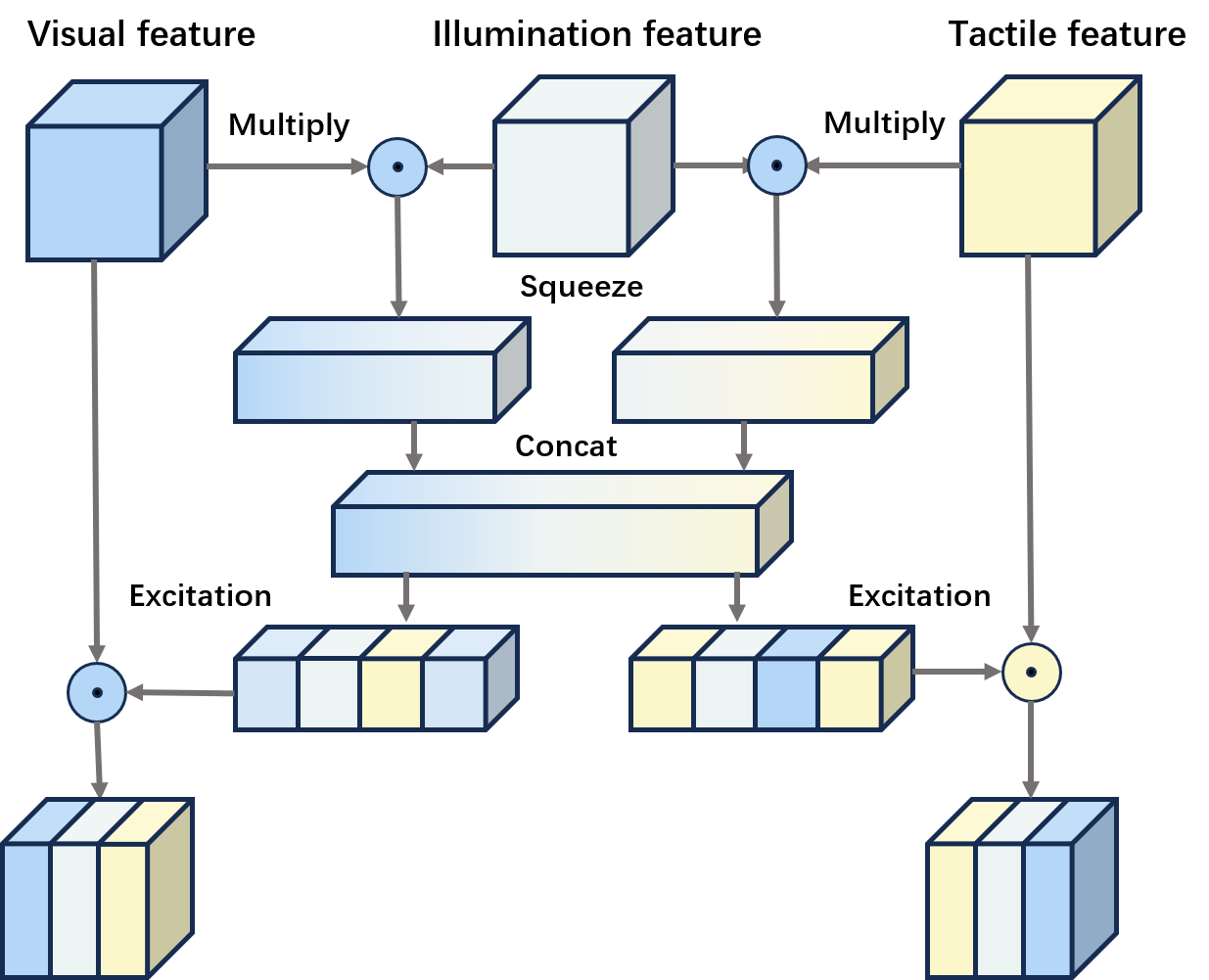}
    \caption{The fusion module with illumination features}
    \label{fig:IMF}
\end{figure}

Then, the illumination features $\mathbf{F}_{i}$ are introduced into the multi-modal fusion module. Inspired by MMTM~\cite{mmtm}, we utilizes the SE mechanism to distribute the weights of both exteroceptive and proprioceptive modalities under varying lighting conditions. The structure of the multi-modal fusion layer with illumination is shown in Fig.~\ref{fig:IMF}.

 Illumination features $\mathbf{F}_{i}$ are multiplied with features of two modalities $\mathbf{F}_{e}^{'l}$ and $1-\mathbf{F}_{i}$ are multiplied with proprioceptive features $\mathbf{F}_{p}^{'l}$. Global average pooling operation is applied to both multiplied features to squeeze the spatial information into the channel descriptors as Eq. ~\ref{equation:combine1} and  Eq. ~\ref{equation:combine2}.
\begin{equation}
\mathbf{S}_{e}^{l}={f}_{sq}\left(\mathbf{F}_{e}^{'l},\mathbf{F}_{i}\right)=\frac{1}{H \times W} \sum_{i=1}^{H} \sum_{j=1}^{W} [F_e^{'l}*F_i](i, j)
\label{equation:combine1}
\end{equation}
\begin{equation}
\mathbf{S}_{p}^{l}={f}_{sq}\left(\mathbf{F}_{p}^{'l},\mathbf{F}_{i}\right)=\frac{1}{H \times W} \sum_{i=1}^{H} \sum_{j=1}^{W} [F_p^{'l}*(1-F_i)](i, j)
\label{equation:combine2}
\end{equation}

Subsequently, the spatial information of each modality are concatenated together and different attention weights of each channel are calculated through two fully connected layers as shown in Eq. \ref{Eq:fc1} and 
 Eq.  \ref{Eq:fc2}.
\begin{equation}
\label{Eq:fc1}
\mathbf{Z}=\mathbf{W}\left[\mathbf{S}_{e}, \mathbf{S}_{p}\right]+b
\end{equation}
\begin{equation}
\label{Eq:fc2}
\mathbf{E}_{e}^{l}=\mathbf{W}_{e} \mathbf{Z}+b_{e}, \quad \mathbf{E}_{p}^{l}=\mathbf{W}_{p} \mathbf{Z}+b_{p}
\end{equation}

The output signals of two modalities $\mathbf{F}_{e}^{l+1}$, $\mathbf{F}_{p}^{l+1}$ of the multi-modal fusion module are generated by a gating mechanism  that re-calibrates features of both modalities with the attention weights in Eq. \ref{Eq:output}.
\begin{equation}
\label{Eq:output}
    \begin{split}
\mathbf{F}_{e}^{l+1} & = \sigma\left(\mathbf{E}_{e}^{l}\right) \odot \mathbf{F}_{e}^{l}\\
\mathbf{F}_{p}^{l+1} & = \sigma\left(\mathbf{E}_{p}^{l}\right) \odot \mathbf{F}_{p}^{l}
    \end{split}
\end{equation}

where $\sigma(\cdot)$ denotes the Sigmoid function and $\odot$ represents the channel-wise product operation. The scaled signals for both modalities are subsequently fed into the next layer for further feature extraction and information fusion. In this paper, we utilize two layers of residual blocks along with the multi-modal fusion layer in the visual-tactile fusion module to amplify the influence of lighting conditions on the fusion process.

\textbf{Road terrain classifier:} In the end, we design the classifiers $f_C$ for exteroceptive and proprioceptive data, separately. Each classifier contains two fully connected layers and a dropout layer in case of overfitting. The average of the each classifier is calculated as the final perception result of the algorithm.
\begin{equation}
    \begin{split}
\mathbf{y}_{e}&={f}^{cla}_{e}\left(\mathbf{F}_{e}\right)\\
\mathbf{y}_{p}&={f}^{cla}_{p}\left(\mathbf{F}_{p}\right)\\
\mathbf{y_{r}}&=(\mathbf{y}_{e}+\mathbf{y}_{p})/2
    \end{split}
\end{equation}
\subsection{Optimization process}

In the training process, the illumination perception sub-network $f_{i}$ is first pre-trained to calculate the illumination features and illumination loss $\mathcal{L}_{i}$. During the training process of the road classifier, the illumination features are input into the multi-modal fusion module. In addition, the illumination loss is also added to the overall loss $\mathcal{L}$ to enhance the influence of lighting conditions on the multi-modal fusion process.
\begin{equation}
\label{Eq:totalloss}
\mathcal{L}=\mathcal{L}_{r}+\lambda \cdot \mathcal{L}_{i}
\end{equation}
The $\lambda$ is a hyperparameter and needs to be fine-tuned and we set $\lambda=1$. The overall training process of the proposed network is demonstrated as follows.

\begin{algorithm} 
\label{algoritm}
	\caption{Training process of the visual-tactile fusion algorithm} 
	\label{alg3} 
    \textbf{Input:} visual and tactile data pairs $\{\mathbf{x}_{v}^{i},\mathbf{x}_t^{i}\}$ and corresponding road labels and illumination labels $\{\mathbf{y}_{cla}^{i},\mathbf{y}_{light}^{i}\}$. Total epochs \emph{Epochs},training batch size  \emph{bs}, learning rate $\ell$.\\
    \textbf{Output:} the illumination perception sub-network $f_{light}$ the feature extractor $f_G$, fusion module $f_F$ and label classifier $f_C$.
    \begin{algorithmic}[1]
        \FOR {$epoch$ in  \emph{Epochs}}
            \FOR{ $batch$ in Batches}
            \STATE Get image data $\mathbf{x}_v^{i}$ and corresponding illumination labels $\mathbf{y}_{light}^{i}$
            \STATE Calculate the predict illumination labels $\hat{\mathbf{y}}_{light}^{i}=f_{light}(\mathbf{x}_v^{i})$ and illumination features $\mathbf{F}_{i}$ 
            \STATE Calculate the illumination loss $\mathcal{L}_{light}(\hat{\mathbf{y}}_{light}^{i},\mathbf{y}_{light}^{i})$
            \STATE Update the parameters of illumination perception sub-network $f_{light}$ by Adam optimizer.   
            \ENDFOR 
        \ENDFOR
        \FOR {$epoch$ in  \emph{Epochs}}
            \FOR{ $batch$ in Batches}
            \STATE Get multimodal data pairs $\{\mathbf{x}_v^{i},\mathbf{x}_t^{i}\}$ and corresponding road labels $\mathbf{y}_{cla}^{i}$
            \STATE Generate the predicted illumination features $\mathbf{F}_{i}$ output from $f_{light}$
            \STATE Calculate the predicted road condition $\hat{\mathbf{y}}_{cla}^{i}=f_G(f_F(f_G(\mathbf{x}_v^{i},\mathbf{x}_t^{i}),\mathbf{F}_{i}))$ 
            \STATE Calculate the classifier loss $\mathcal{L}_{cla}(\hat{\mathbf{y}}_{cla}^{i},\mathbf{y}_{cla}^{i})$ and the final loss $\mathcal{L}=\mathcal{L}_{cla}+\lambda \cdot \mathcal{L}_{light}$
            \STATE Update the parameters of the visual-tactile fusion network by Adam optimizer.        
            \ENDFOR 
        \ENDFOR    

	\end{algorithmic} 
\end{algorithm}

\section{EXPERIMENTAL RESULTS and ANALYSIS}
\label{Sec:results}
\subsection{Dataset}

To validate the effectiveness of the proposed method, we constructed two multi-modal datasets that incorporate various types of sensors and operating conditions. These datasets allows us to assess the proposed algorithm's performance under different environmental conditions and operational scenarios. A detailed description is provided below.

\subsection{Dataset1: contains acceleration and images}

\begin{figure}[htp]
    \centering
    \includegraphics[width=8cm]{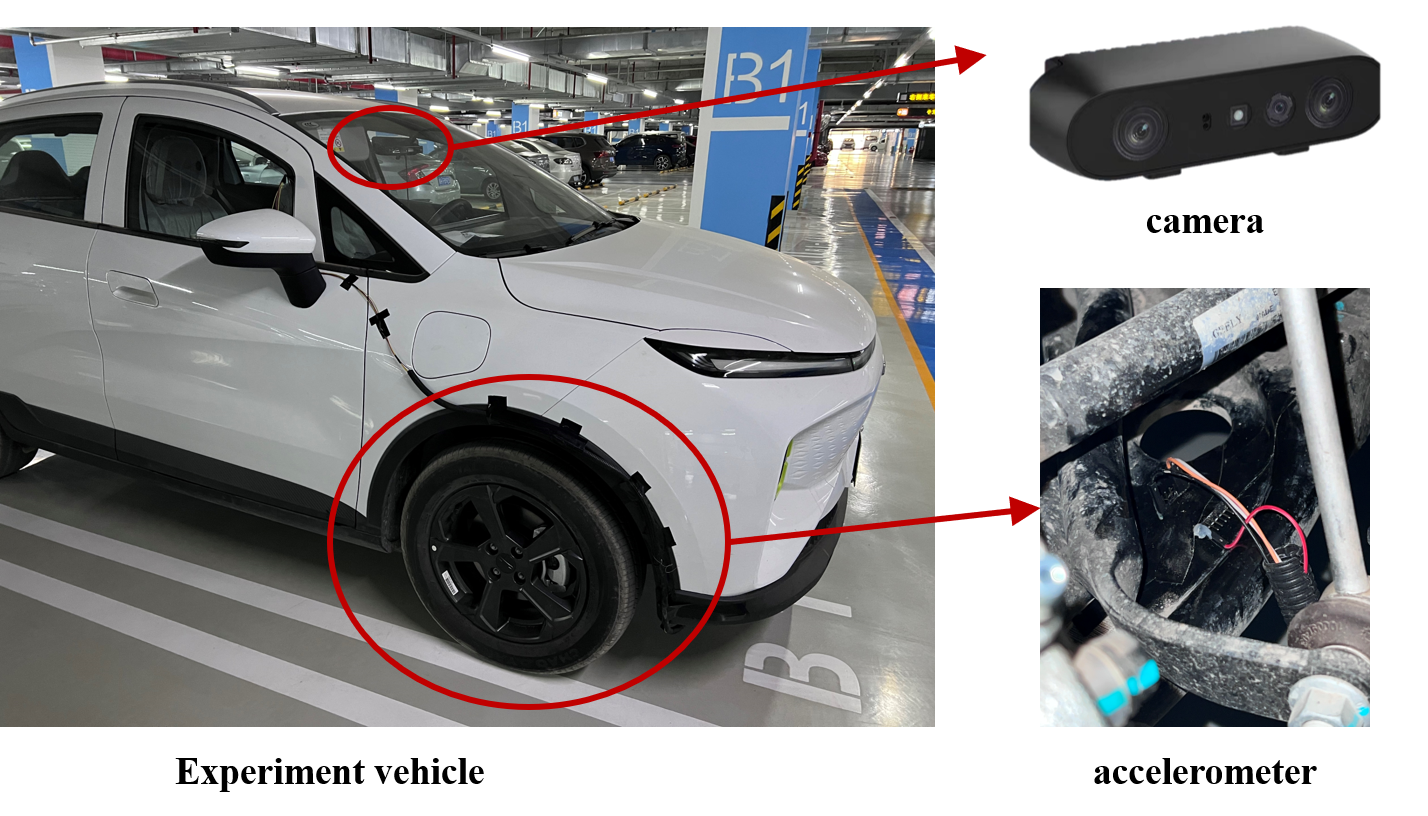}
    \caption{The experiment vehicle and the sensors installation.}
    \label{fig:vehile1}
\end{figure}

For dataset1, we selected a Vette WT931 accelerometer as the proprioception sensor, mounting it on the suspension of the vehicle’s right front wheel. An IMX307 binocular camera was chosen as the exteroception sensor, installed on the front windshield. The sampling rates of the accelerometer and camera are set to 500 Hz and 60 fps, respectively. These sensors were mounted on a Geely Geometry E passenger vehicle shown in Fig~\ref{fig:vehile1}, with data collected via a connected laptop.

This dataset aims to validate the proposed algorithm's recognition accuracy under different lighting conditions and vehicle speeds. Although only three types of road surfaces: asphalt, gravel, and concrete were selected, we incorporated a comprehensive range of lighting conditions: noon, dusk, and night. Additionally, to comprehensively compare the impact of vehicle speed on recognition performance, we maintained the same speeds of 10 km/h, 20 km/h, and 30 km/h across different road surfaces, since it can be hazardous when driving at a higher speed on gravel roads. These conditions were chosen to control variables and comprehensively evaluate recognition performance in different operational scenarios and lighting environments. 

For acceleration data, we generate the corresponding spectrogram through a sliding window and the wavelet transform, which are taken as the input of the fusion multi-modal network. First, a sliding window is applied to each acceleration sequence $[a_i]_{i=0,1,\cdots,L}$ to generate the acceleration data $A^{i}_{l}$ corresponding to single image:
\begin{equation} 
\begin{aligned} 
A^{i}_{l}=\left[a_{i}, a_{i+1}, a_{i+2},\ldots a_{i+l}\right] \\
\quad i=0, \Delta n, 2 \Delta n, \cdots, (L / / \Delta n)*\Delta n
\end{aligned}
\end{equation}
where $l$ is the length of single acceleration array and here we set $l=500$. $L$ is the total length of each raw acceleration sequence.$\Delta n$ is the moving step for sliding windows. The image data are then selected according to time index synchronously .  
Further, we use  Continous Wavelet Transform(CWT) to convert the original one-dimensional data into 256x256 spectrogram, whose formula is shown as  Eq.~\ref{equation:cwt}:
\begin{equation} 
W(a, b)=\int_{-\infty}^{\infty} A^{i}_{l} \cdot \psi\left(\frac{t-b}{a}\right) dt
\label{equation:cwt}
\end{equation}
 where $W(a,b)$ is the coefficient of wavelet spectrogram, $\psi(t)$ is the wavelet basis function and we choose cgau8 as the basis function. $a$ is the scale parameter and $b$ is the translation parameter ~\cite{cwt}. The generated wavelet spectrogram is represented as $\mathbf{x}_{p}^{i}$ and images are represented as $\mathbf{x}_{e}^{i}$.
 \begin{figure}[htp]
    \centering
\includegraphics[width=7cm]{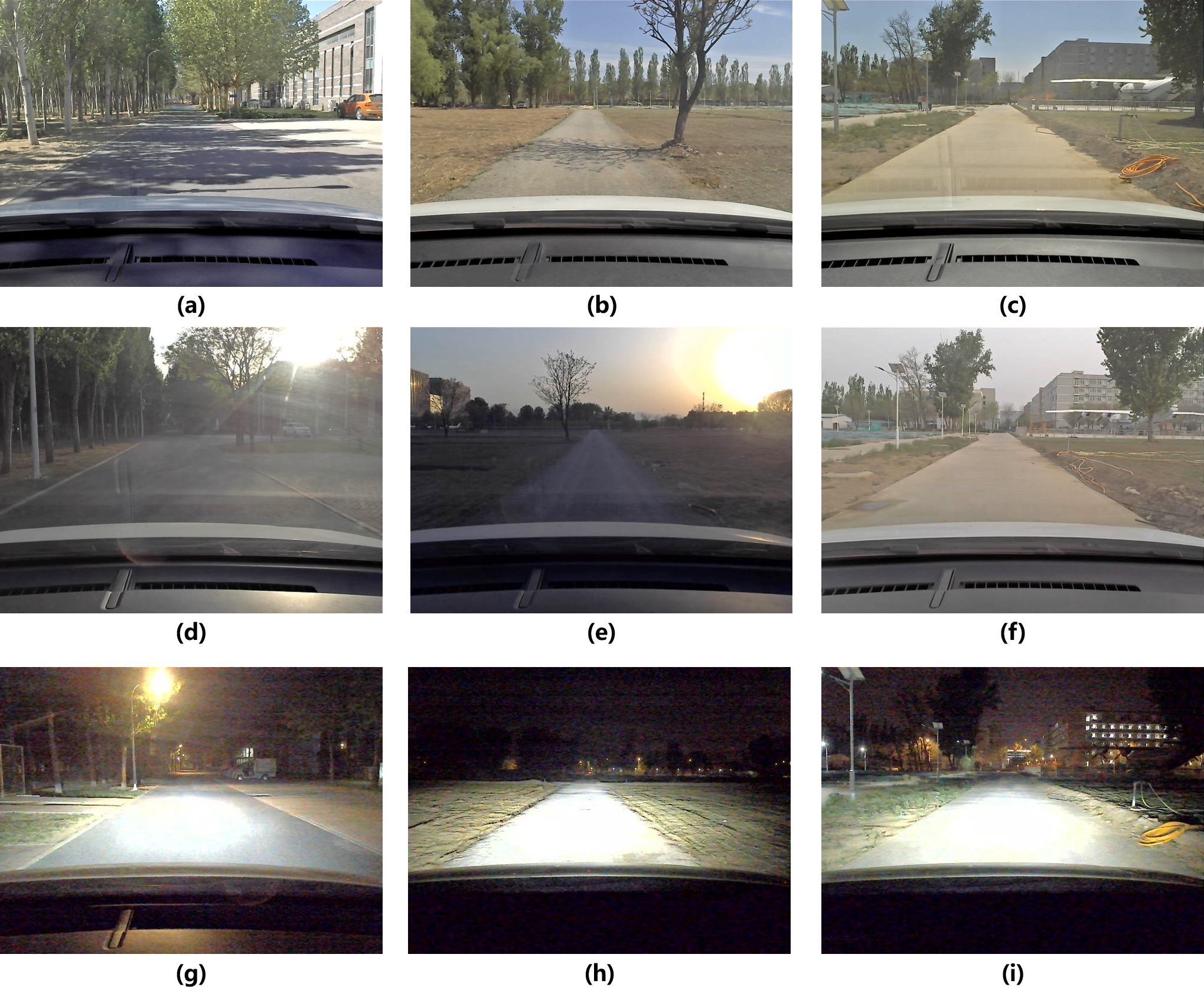}
\caption{The raw acceleration data under different illumination conditions:(a),(d),(g): asphalt at noon, dusk and night; (b),(e),(h): gravel at noon, dusk and night; (c),(f),(i): cement at noon, dusk and night.}
\label{fig:road_img_xiaonei}
\end{figure}

\begin{figure}[htbp]
\centering
\begin{minipage}[t]{0.47\columnwidth}
\centering
\includegraphics[width=\textwidth]{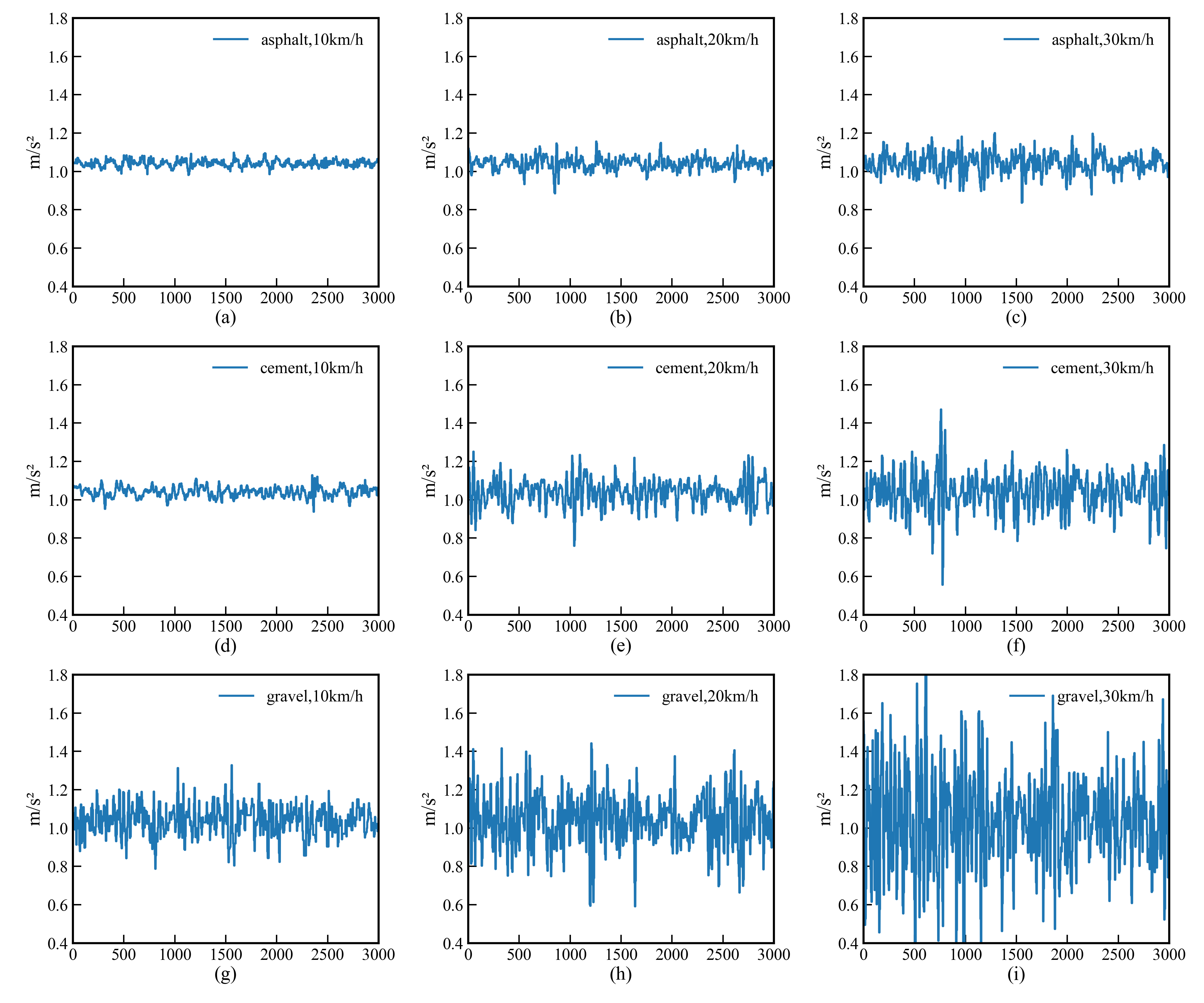}
\caption{The road images under different illumination conditions:(a),(d),(g): asphalt at 10, 20 and 30km/h; (b),(e),(h): gravel at 10, 20 and 30km/h; (c),(f),(i): cement at 10, 20 and 30km/h.}
\label{fig:road_acc_xiaonei}
\end{minipage}
\hspace{0.04\textwidth} 
\begin{minipage}[t]{0.47\columnwidth}
\centering
\includegraphics[width=\textwidth]{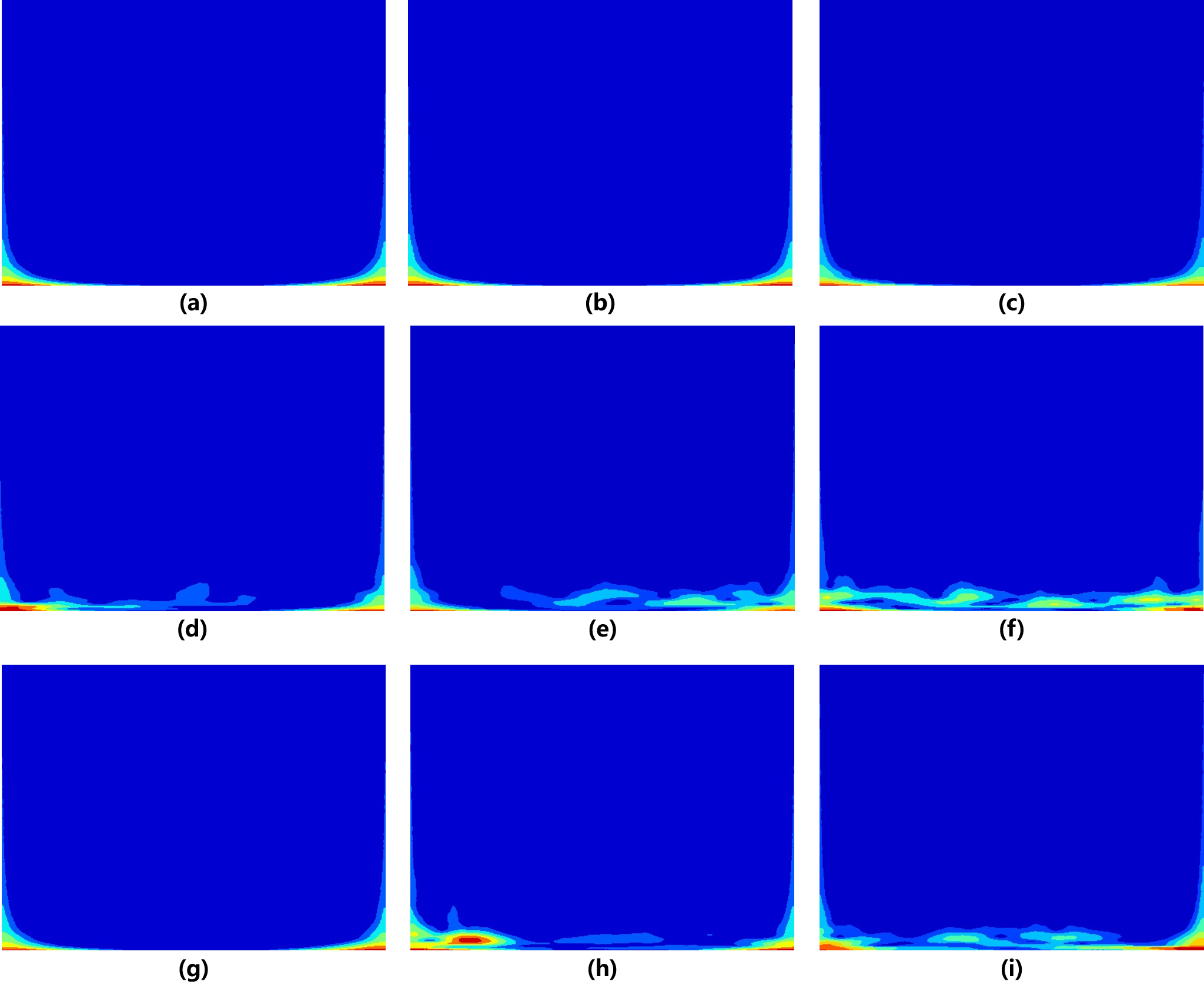}
\caption{The cwt spectrogram of acceleration under differernt working conditions:(a),(d),(g): asphalt at 10, 20 and 30km/h; (b),(e),(h): gravel at 10, 20 and 30km/h; (c),(f),(i): cement at 10, 20 and 30km/h.}
\label{fig:road_cwt_xiaonei}
\end{minipage}
\end{figure}

The visual data under varying lighting conditions is illustrated in the Fig \ref{fig:road_img_xiaonei} , the raw acceleration data and corresponding spectrogram images at different speeds are also shown in Fig \ref{fig:road_acc_xiaonei} and \ref{fig:road_cwt_xiaonei}. It is observed that cwt spectrograms effectively extract features and standardize the proprioceptive modality data to the same format as the exteroceptive data, which is convenient for the fusion process. Finally, we take spectrograms as proprioceptive input. The details of this dataset are demonstrated in \ref{Table:dataset1}.



\begin{table}

\scriptsize
\centering  
\caption{Details of dataset1: acceleration and images }
\begin{tabular}{ccccc}
\hline
\textbf{Road type}                    & \textbf{light condition} & \textbf{10km/h} & \textbf{20km/h} & \textbf{30km/h} \\ \hline
\multirow{3}{*}{\textbf{gravel}}  & noon             & 579    & 309    & 189    \\
                         & dusk             & 469    & 314    & 213    \\
                         & night            & 623    & 337    & 208    \\ \hline
\multirow{3}{*}{\textbf{asphalt}} & noon             & 755    & 410    & 246    \\
                         & dusk             & 730    & 334    & 240    \\
                         & night            & 694    & 322    & 284    \\ \hline
\multirow{3}{*}{\textbf{cement}}  & noon             & 286    & 150    & 91     \\
                         & dusk             & 350    & 162    & 93     \\
                         & night            & 309    & 145    & 88     \\ \hline
\end{tabular}
\label{Table:dataset1}
\end{table}
\subsection{Dataset2: contained intelligent tires data and images}
\begin{figure}[htp]
    \centering
    \includegraphics[width=10cm]{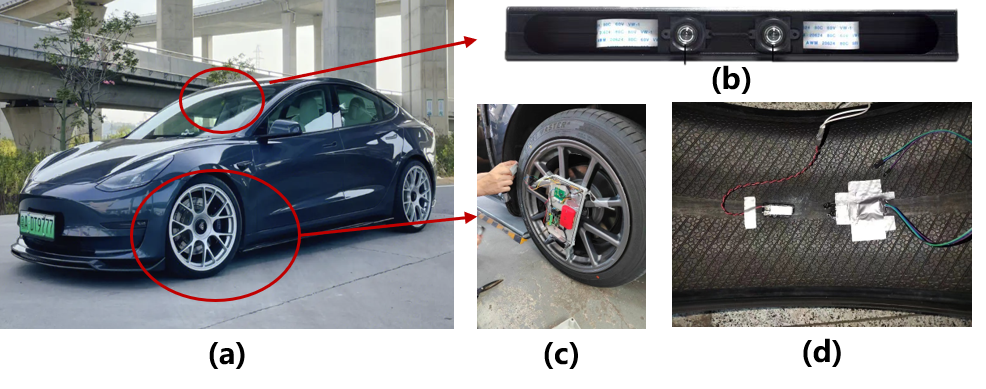}
    \caption{The multi-modal perception system equipped with intelligent tires and a camera:(a) the experiment vehicle; (b) the binocular camera; (c) the intelligent tire system; (d) the PVDF sensor.}
    \label{fig:tesla}
\end{figure}

For dataset2, we developed an intelligent tire system as the proprioception sensor. An DT1-028K PVDF sensor was adhenced to the inner wall of the tire to collect kinematic information. We utilized a Raspberry Pi along with an AD acquisition module ADS 1263 to collect signals in real-time, with wireless communication between the intelligent tire system and a computer. An IMX307 binocular camera, as the exteroception sensor, was similarly mounted on the front windshield. The sampling rates for the intelligent tire and camera were set at 1100 Hz and 60 fps, respectively. These sensors were installed on a Tesla Model 3 shown as \ref{fig:tesla}.

\begin{figure}[htp]
    \centering
    \includegraphics[width=12cm]{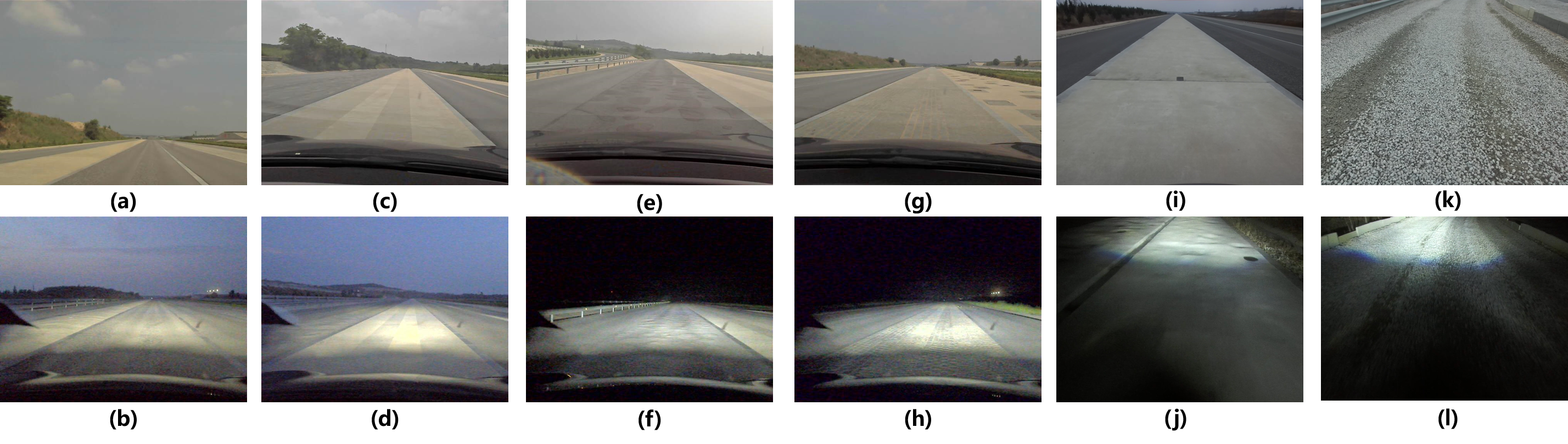}
    \caption{images of different road types under daytime and night: (a)-(c):asphalt road at daytime, night and cwt spectrogram;
    (d)-(f):cecment road at daytime, night and  cwt spectrogram;    (g)-(i):patched asphalt at daytime, night and cwt spectrogram;   (j)-(i):brick road at daytime, night and cwt spectrogram;     (m)-(o):irregular concrete at daytime, night and cwt spectrogram;  (p)-(r):gravel at daytime, night and cwt spectrogram.}
    \label{fig:shiyanchang}
\end{figure}

\begin{figure}[htp]
    \centering
    \includegraphics[width=12cm]{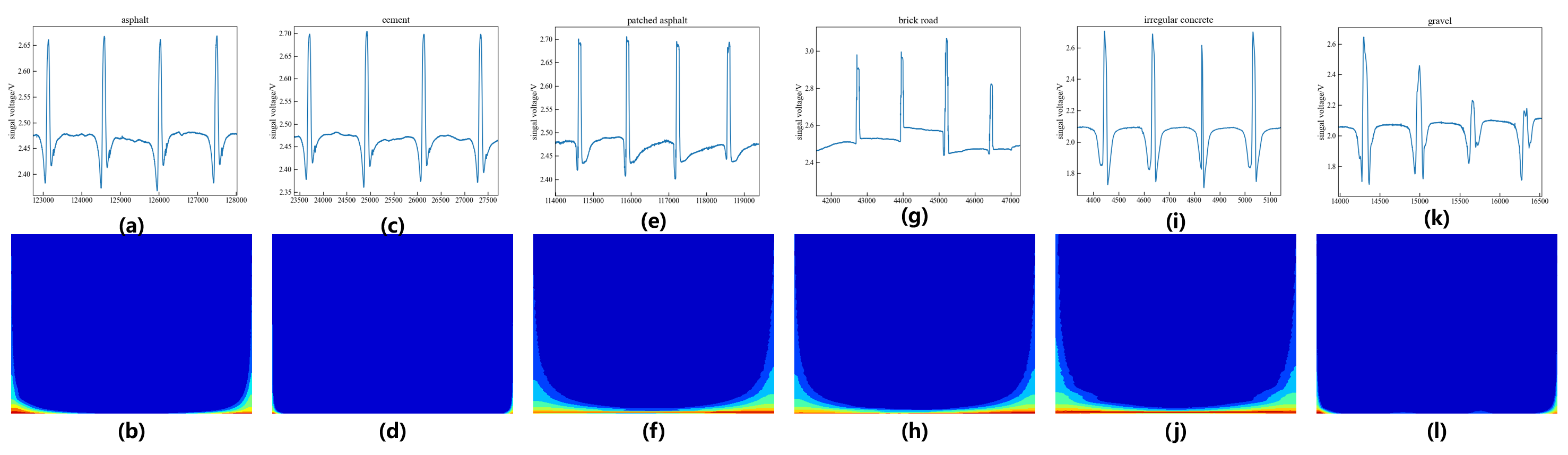}
    \caption{raw intelligent tire data and cwt spectrograms of different road types : (a)-(b):asphalt road;
    (c)-(d):cecment road;    (e)-(f):patched asphalt;   (g)-(h):brick road;     (i)-(j):irregular concrete;  (k)-(l):gravel.}
    \label{fig:shiyanchang_pvdf}
\end{figure}

The dataset2 focuses on a wider variety of road surfaces and vehicle speed settings that are closer to real-world conditions. Six types of road surfaces are included: asphalt, concrete, patched asphalt, brick road, irregular concrete, and gravel. Lighting conditions included both day and night, with speeds ranging from 10 to 80 km/h. The different road images are shown in Fig.\ref{fig:shiyanchang}. Also, the corresponding proprioceptive data and cwt spetroframs are shown in Fig.\ref{fig:shiyanchang_pvdf}. Same as dataset1, iamges and spectrograms are input into the multi-modal fusion network.

We use periodic signal segmentation and wavelet transform to generate the corresponding spectrogram.We identify the peak corresponding to each cycle, then extract the data between adjacent peaks as the data for one full tire rotation. Similarly, the corresponding image data is aligned based on the time index. Furthermore, the wavelet transform same as Dataset1 is applied to the periodic data of the intelligent tire to obtain its spectrogram. Finally, the spectrogram of the intelligent tire is matched with the image data, generating a set of multi-modal data pairs $\mathbf{x}_{p}^{i}$ and $\mathbf{x}_{e}^{i}$.  The details of this dataset are demonstrated in Table\ref{Table:pvdfdata}.

\begin{table}
\scriptsize
\centering  
\caption{Details of dataset2: intelligent tires and images }

\begin{tabular}{cccccc}
\hline
\textbf{Road type}                           & \textbf{light condition} & \textbf{10km/h} & \textbf{30km/h} & \textbf{50km/h} & \textbf{80km/h} \\ \hline
\multirow{2}{*}{\textbf{asphalt}}            & day                      & 283             & 293             & 276             & 286             \\
 & night                    & 151             & 250             & 226             & 162             \\ \hline
\multirow{2}{*}{\textbf{cement}}             & day                      & 157             & 125             & 198             & 148             \\
& night                    & 141             & 141             & 148             & -               \\ \hline
\multirow{2}{*}{\textbf{patched asphalt}}    & day                      & 81              & 88              & -               & -               \\
& night                    & 81              & 88              & -               & -               \\ \hline
\multirow{2}{*}{\textbf{brick road}}         & day                      & 138             & 59              & -               & -               \\
 & night                    & 47              & 88              & -               & -               \\ \hline
\multirow{2}{*}{\textbf{irregular concrete}} & day                      & 45              & 196             & -               & -               \\
& night                    & 148             & 196             & -               & -               \\ \hline
\multirow{2}{*}{\textbf{gravel}}             & day                      & 74              & 107             & -               & -               \\
 & night                    & -               & 86              & -               & -               \\ \hline
\end{tabular}
\label{Table:pvdfdata}

\end{table}

\subsubsection{Experiment settings}

During the training process, the Adam optimizer is utilized with a weight decay of $5\times 10^{-4}$. Learning rate is initialized to $8\times 10^{-4}$ and scheduled using  \emph{lr scheduler.ReduceLROnPlateau} method. Batch size is set to 32 and the number of epochs is 100. All experiments are conducted in the NVIDIA GeForce RTX 4090.

\subsection{Experiment results}

\subsubsection{Compared with baseline methods}

To verify the effectiveness of the proposed algorithm, the road recognition results of IMF is compared with those from other baseline methods. Six types of channel-exchange based fusion methods  are selected as baselines: MMTM~\cite{mmtm}, CEN~\cite{CEN}, EIP~\cite{EIP} take CNN as backbones, TKF~\cite{TKF}, MFT~\cite{MFT}, MBT ~\cite{MBT},DSF~\cite{DSF}, MMSF~\cite{MMSF} take Transformer as backbones. We also design three types of aggregation-based fusion methods with CNN as backbones, which are early-fusion, mid-fusion and late-fusion, respectively. 

The comparative analysis between the proposed method and other baseline methods are shown in Table~\ref{Table:baselinecompare} and Table~\ref{Table:baselinepvdf}. For dataset1 ,we focus on recognition accuracy across different light conditions and speeds, and thus, only accuracy are demonstrated. In Table~\ref{Table:baselinecompare}, the highest accuracy under each working condition is highlighted in bold red and the second highest accuracy  is marked in bold blue. For dataset2, we compare different metrics in order to analyze the influence of light conditions on road recognition performance, with the highest values similarly highlighted in bold red.


From Table~\ref{Table:baselinecompare}, it is evident that variations in lighting conditions and vehicle speed have a significant impact on the recognition results. The proposed IMF achieves the highest recognition accuracy in five out of nine conditions, outperforming other baseline methods. This demonstrates that IMF is capable of effectively recognizing road surfaces across different lighting conditions and vehicle speeds, indicating its robustness in varying operational environments.


From Table~\ref{Table:baselinepvdf} for dataset2, we observe that while IMF performs relatively poorly in terms of precision, recall, and F1 score during daytime conditions compared to other baselines, it achieves the highest overall recognition accuracy during the day. Moreover, IMF significantly outperforms other methods in nighttime conditions, obtaining the best recognition results across all four evaluation metrics: precision, recall, F1 score, and accuracy. In summary, the comparative analysis indicates that IMF is capable of achieving satisfactory recognition performance in both daytime and nighttime conditions.

In conclusion, the advantages of IMF lie in its consistent high performance across different light conditions and speeds, suggesting that its fusion approach is better at capturing road features compared to traditional baselines.

\begin{table*}[]
\centering
\scriptsize
\caption{accuracy comparison with baselines for dataset1}

\begin{tabular}{cccccccccc}
\hline
\textbf{light condition} & \multicolumn{3}{c}{\textbf{noon}}                                                                                        & \multicolumn{3}{c}{\textbf{dusk}}                                                                                        & \multicolumn{3}{c}{\textbf{night}}                                                                                       \\ \hline
\textbf{speed(km/h)}     & \textbf{10}                            & \textbf{20}                            & \textbf{30}                            & \textbf{10}                            & \textbf{20}                            & \textbf{30}                            & \textbf{10}                            & \textbf{20}                            & \textbf{30}                            \\ \hline
\textbf{MMTM}            & {\color[HTML]{CB0000} \textbf{0.8906}} & {\color[HTML]{3166FF} \textbf{0.8594}} & {\color[HTML]{333333} 0.8594}          & {\color[HTML]{3166FF} \textbf{0.9219}} & {\color[HTML]{000000} 0.9062}          & {\color[HTML]{3166FF} \textbf{0.8594}} & {\color[HTML]{CB0000} \textbf{0.875}}  & {\color[HTML]{CB0000} \textbf{0.9219}} & 0.8672                                 \\
\textbf{CEN}             & 0.8125                                 & 0.8125                                 & {\color[HTML]{3166FF} \textbf{0.9375}} & 0.8438                                 & 0.9062                                 & 0.8281                                 & 0.7656                                 & 0.8906                                 & {\color[HTML]{3166FF} \textbf{0.9062}} \\
\textbf{EIP}             & 0.7969                                 & {\color[HTML]{333333} 0.7969}          & {\color[HTML]{333333} 0.9062}          & {\color[HTML]{333333} 0.8125}          & 0.9062                                 & 0.8438                                 & {\color[HTML]{3166FF} \textbf{0.8125}} & 0.8438                                 & 0.8594                                 \\ \cline{1-1}
\textbf{mbt}             & 0.4688                                 & 0.4219                                 & 0.5469                                 & 0.5469                                 & 0.5625                                 & 0.3594                                 & 0.5469                                 & 0.4375                                 & 0.5234                                 \\
\textbf{MFT}             & {\color[HTML]{3166FF} \textbf{0.8438}} & {\color[HTML]{000000} 0.8438}          & 0.8906                                 & {\color[HTML]{000000} 0.9062}          & {\color[HTML]{343434} 0.9219}          & {\color[HTML]{333333} 0.8438}          & 0.75                                   & {\color[HTML]{3166FF} \textbf{0.9062}} & {\color[HTML]{CB0000} \textbf{0.9375}} \\
\textbf{TKF}             & 0.7917                                 & 0.8333                                 & 0.7812                                 & 0.8125                                 & 0.8333                                 & 0.7812                                 & 0.7708                                 & 0.8125                                 & 0.8203                                 \\
\textbf{DSF}             & 0.8594                                 & 0.6875                                 & 0.8438                                 & 0.7031                                 & 0.7344                                 & 0.7812                                 & 0.5156                                 & 0.75                                   & 0.6875                                 \\
\textbf{MMSF}            & 0.8438                                 & 0.7812                                 & 0.9375                                 & 0.8906                                 & 0.9375                                 & 0.8594                                 & 0.7812                                 & 0.9219                                 & 0.8438                                 \\ \cline{1-1}
\textbf{early fusion}    & 0.8281                                 & 0.75                                   & {\color[HTML]{CB0000} \textbf{0.9531}} & {\color[HTML]{333333} 0.875}           & 0.9062                                 & 0.8438                                 & 0.75                                   & 0.875                                  & {\color[HTML]{3166FF} \textbf{0.9062}} \\
\textbf{middle fusion}   & {\color[HTML]{CB0000} \textbf{0.8906}} & {\color[HTML]{333333} 0.8281}          & 0.9062                                 & 0.8906                                 & 0.9062                                 & {\color[HTML]{3166FF} \textbf{0.8594}} & {\color[HTML]{3166FF} \textbf{0.8125}} & {\color[HTML]{CB0000} \textbf{0.9219}} & 0.8672                                 \\
\textbf{late fusion}     & {\color[HTML]{333333} 0.8594}          & {\color[HTML]{CB0000} \textbf{0.8906}} & {\color[HTML]{333333} 0.9219}          & {\color[HTML]{333333} 0.875}           & {\color[HTML]{3166FF} \textbf{0.9375}} & 0.8438                                 & {\color[HTML]{333333} 0.7812}          & 0.8906                                 & 0.8984                                 \\ \hline
\textbf{IMF}             & {\color[HTML]{CB0000} \textbf{0.9219}} & {\color[HTML]{333333} 0.8438}          & {\color[HTML]{CB0000} \textbf{0.9531}} & {\color[HTML]{CB0000} \textbf{0.9844}} & {\color[HTML]{CB0000} \textbf{0.9844}} & {\color[HTML]{CB0000} \textbf{0.875}}  & {\color[HTML]{CB0000} \textbf{0.875}}  & {\color[HTML]{000000} 0.8906}          & {\color[HTML]{333333} 0.8984}          \\ \hline
\end{tabular}
\label{Table:baselinecompare}
\end{table*}

\begin{table}
\centering
\scriptsize
\caption{different metrics comparison with baselines for dataset2}
\begin{tabular}{ccccccccc}
\hline
\textbf{light condition} & \multicolumn{4}{c}{\textbf{day}}                                                                                                                                  & \multicolumn{4}{c}{\textbf{light}}                                                                                                                                \\ \hline
\textbf{speed(km/h)}     & \textbf{precision}                     & \textbf{recall}                        & \textbf{f1}                            & \textbf{acc}                           & \textbf{precision}                     & \textbf{recall}                        & \textbf{f1}                            & \textbf{acc}                           \\ \hline
\textbf{MMTM}            & {\color[HTML]{3166FF} \textbf{0.9629}} & {\color[HTML]{CB0000} \textbf{0.9871}} & {\color[HTML]{CB0000} \textbf{0.9740}} & {\color[HTML]{3166FF} \textbf{0.9757}} & 0.9139                                 & 0.9643                                 & 0.9333                                 & 0.9594                                 \\
\textbf{CEN}             & {\color[HTML]{CB0000} \textbf{0.9644}} & {\color[HTML]{3166FF} \textbf{0.9705}} & {\color[HTML]{3166FF} \textbf{0.9672}} & {\color[HTML]{000000} 0.9740}          & {\color[HTML]{3166FF} \textbf{0.9588}} & {\color[HTML]{3166FF} \textbf{0.9786}} & {\color[HTML]{3166FF} \textbf{0.9673}} & {\color[HTML]{3166FF} \textbf{0.9688}} \\
\textbf{EIP}             & 0.9554                                 & 0.9502                                 & 0.9527                                 & 0.9705                                 & 0.9307                                 & 0.9367                                 & 0.9307                                 & 0.9531                                 \\ \hline
\textbf{mbt}             & 0.1630                                 & 0.0834                                 & 0.0981                                 & 0.3646                                 & 0.1668                                 & 0.0959                                 & 0.1064                                 & 0.3875                                 \\
\textbf{MFT}             & 0.9126                                 & 0.9375                                 & 0.9228                                 & 0.9288                                 & 0.8345                                 & 0.8463                                 & 0.8361                                 & 0.8844                                 \\
\textbf{TKF}             & {\color[HTML]{000000} 0.8333}          & {\color[HTML]{000000} 0.7777}          & {\color[HTML]{000000} 0.8}             & 0.9253                                 & 0.7094                                 & 0.7583                                 & 0.7277                                 & 0.9148                                 \\
\textbf{DSF}             & \multicolumn{1}{l}{0.7983}             & \multicolumn{1}{l}{0.8809}             & \multicolumn{1}{l}{0.7605}             & \multicolumn{1}{l}{0.8681}             & \multicolumn{1}{l}{0.7035}             & \multicolumn{1}{l}{0.5648}             & \multicolumn{1}{l}{0.6027}             & \multicolumn{1}{l}{0.7219}             \\
\textbf{MMSF}            & \multicolumn{1}{l}{0.6582}             & \multicolumn{1}{l}{0.6645}             & \multicolumn{1}{l}{0.6545}             & \multicolumn{1}{l}{0.7378}             & \multicolumn{1}{l}{0.6619}             & \multicolumn{1}{l}{0.6848}             & \multicolumn{1}{l}{0.6412}             & \multicolumn{1}{l}{0.7375}             \\ \hline
\textbf{early fusion}    & 0.9349                                 & 0.9575                                 & 0.9441                                 & 0.9705                                 & 0.9040                                 & 0.934                                  & 0.9144                                 & 0.9500                                 \\
\textbf{middle fusion}   & 0.9147                                 & 0.9358                                 & 0.9229                                 & 0.9566                                 & 0.8516                                 & 0.9228                                 & 0.8684                                 & 0.9187                                 \\
\textbf{late fusion}     & 0.9391                                 & 0.9637                                 & 0.9494                                 & 0.9670                                 & 0.9140                                 & 0.9652                                 & 0.9341                                 & 0.9531                                 \\ \hline
\textbf{IMF}             & 0.9580                                 & 0.9622                                 & 0.9601                                 & {\color[HTML]{CB0000} \textbf{0.9774}} & {\color[HTML]{CB0000} \textbf{0.9607}} & {\color[HTML]{CB0000} \textbf{0.9811}} & {\color[HTML]{CB0000} \textbf{0.9697}} & {\color[HTML]{CB0000} \textbf{0.9781}} \\ \hline
\end{tabular}
\label{Table:baselinepvdf}
\end{table}



\subsubsection{Compared with single-modal data}
\begin{figure}[htp]
    \centering
    \includegraphics[width=12cm]{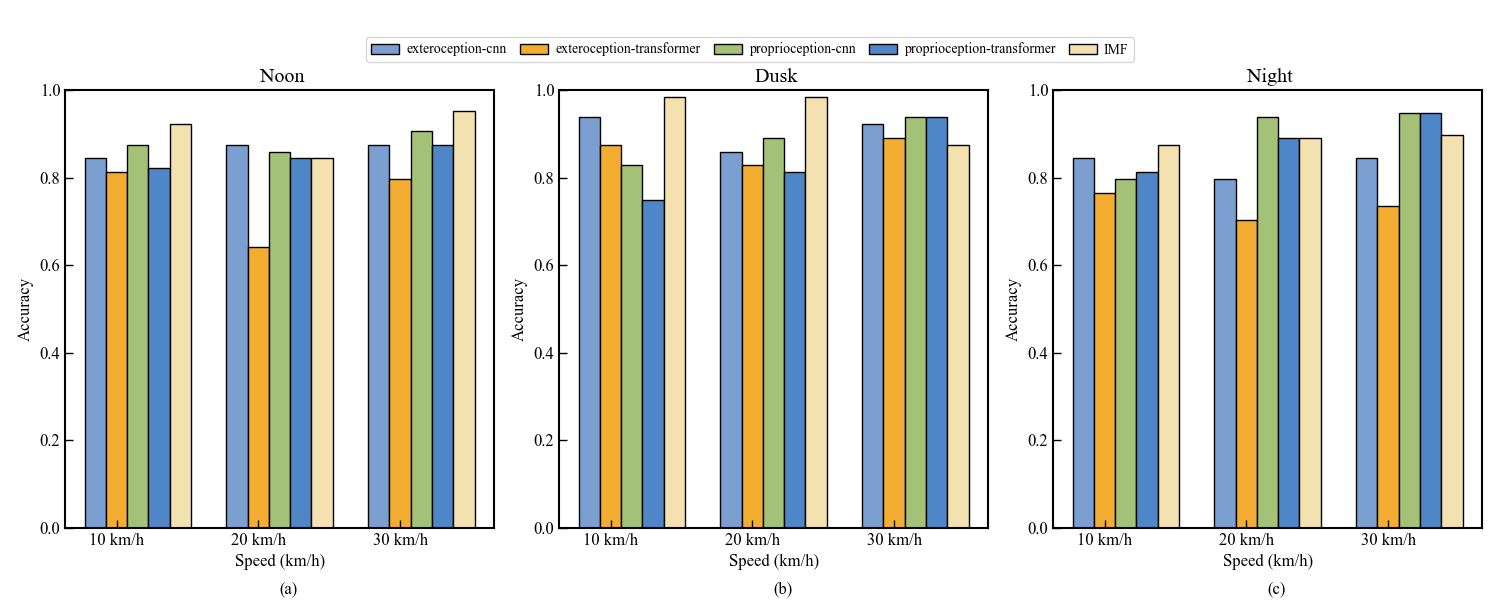}
    \caption{ accuracy comparison with methods based on single modality on dataset1.}
    \label{fig:single_modal_acc}
\end{figure}

\begin{figure}[htp]
    \centering
    \includegraphics[width=10.5cm]{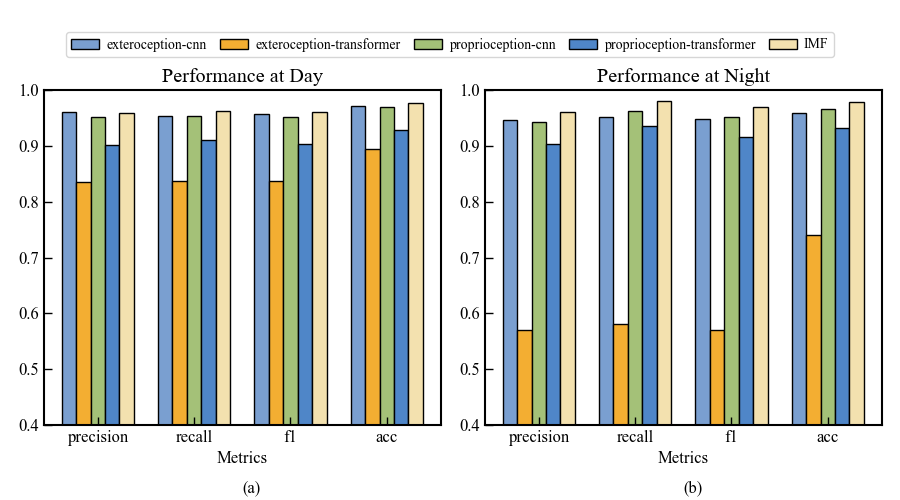}
    \caption{ accuracy comparison with methods based on single modality on dataset2.}
    \label{fig:single_modal_pvdf}
\end{figure}

In order to verify the necessity of multi-modal fusion method for road perception of AVs, we also compared method IMF against terrain perception algorithms utilizing either a single proprioceptive or exteroceptive modality. Both CNN and Transformer were used as backbones for each modality. For Dataset 1,  Fig.~\ref{fig:single_modal_acc} presents recognition accuracy under different lighting conditions and speeds. For Dataset 2, Fig.~\ref{fig:single_modal_pvdf} demonstrates various evaluation metrics for both daytime and nighttime.

In Fig.~\ref{fig:single_modal_acc}, which compares road recognition accuracy for dataset1, the multi-modal fusion algorithm, IMF, outperforms single modality perception methods in five out of nine conditions. For instance, IMF performs better than single modality methods at 10km/h and 20km/h at dusk. By fusing both proprioception and exteroception modalities, the model effectively leverages complementary information, achieving superior performance across various lighting conditions and speeds.
Similarly, in Fig.~\ref{fig:single_modal_pvdf} for dataset2,  the multi-modal fusion approach consistently outperforms across all metrics—precision, recall, F1-score, and accuracy under varying lighting conditions. For instance, IMF achieves the highest accuracy  at night, significantly surpassing the highest accuracy of the proprioception method and the exteroception method. This consistent advantage highlights how combining proprioceptive and exteroceptive inputs enables the algorithm to capture more comprehensive road features, thereby enhancing recognition accuracy across different scenarios.

In summary, the multi-modal fusion method IMF offers significant advantages by integrating richer and more diverse sensory data, resulting in more accurate and robust road recognition compared to single-modality approaches.

\subsubsection{Ablation study}
 We further conduct the ablation study to evaluate the effectiveness of different modules of IMF. We remove the illumination loss and the illumination perception sub-network, respectively. The recognition accuracy of all algorithms across varying working conditions of Dataset1 are shown in Table~\ref{Table:ablation_acc}, while recognition metrics for both day and night of Dataset2 are demonstrated in Table~\ref{Table:ablation_pvdf}.
 In both Tables, the highest accuracy for each condition is highlighted in bold red.

 In Table~\ref{Table:ablation_acc}, for dataset1, the IMF method consistently performs better across different speed and lighting conditions,particularly  during nighttime.  For instance, under the condition of 20 km/h at dusk, IMF achieves an accuracy of 0.9844, significantly outperforming the "no lighting perception loss" setting (accuracy of 0.8594) and showing comparable performance to the "no lighting condition perception" setting (accuracy of 0.9375). This indicates that IMF's ability to handle various lighting conditions, including challenging scenarios like night-time driving, is superior, enabling more accurate road recognition results.

In Table~\ref{Table:ablation_pvdf} for dataset2, IMF also demonstrates superior recognition results across precision, recall, F1-score, and accuracy metrics compared to other module settings. Under daytime conditions, IMF achieves the highest road types recognition accuracy. In addition, for nighttime conditions, IMF outperforms the other two module settings across all four metrics.  This demonstrates that IMF's multi-modal design effectively tackles varying lighting and environmental conditions, leading to improved road recognition accuracy in diverse scenarios.

To sum up, IMF's ability to incorporate and manage both illumination and road types perception makes it more robust and reliable compared to other module settings, as evidenced by its superior performance across different conditions and datasets.

\begin{table}
\scriptsize
\caption{accuracy comparison with other different module settings for dataset1}
\begin{tabular}{clllllllll}
\hline
\textbf{light condition}                                                    & \multicolumn{3}{c}{\textbf{noon}}                                                                                        & \multicolumn{3}{c}{\textbf{dusk}}                                                                                        & \multicolumn{3}{c}{\textbf{night}}                                                                                       \\ \hline
\textbf{speed(km/h)}                                                        & \multicolumn{1}{c}{\textbf{10}}        & \multicolumn{1}{c}{\textbf{20}}        & \multicolumn{1}{c}{\textbf{30}}        & \multicolumn{1}{c}{\textbf{10}}        & \multicolumn{1}{c}{\textbf{20}}        & \multicolumn{1}{c}{\textbf{30}}        & \multicolumn{1}{c}{\textbf{10}}        & \multicolumn{1}{c}{\textbf{20}}        & \multicolumn{1}{c}{\textbf{30}}        \\ \hline
\begin{tabular}[c]{@{}c@{}}no lighting\\  perception loss\end{tabular}      & 0.9062                                 & {\color[HTML]{CB0000} \textbf{0.8594}} & 0.8281                                 & 0.8906                                 & 0.8594                                 & 0.8906                                 & 0.8125                                 & 0.9375                                 & 0.8906                                 \\
\begin{tabular}[c]{@{}c@{}}no lighting \\ condition perception\end{tabular} & 0.8281                                 & 0.8438                                 & 0.8438                                 & 0.9062                                 & 0.9375                                 & {\color[HTML]{CB0000} \textbf{0.9062}} & {\color[HTML]{CB0000} \textbf{0.8906}} & {\color[HTML]{CB0000} \textbf{0.9531}} & 0.8906                                 \\
IMF                                                                         & {\color[HTML]{CB0000} \textbf{0.9219}} & 0.8438                                 & {\color[HTML]{CB0000} \textbf{0.9531}} & {\color[HTML]{CB0000} \textbf{0.9844}} & {\color[HTML]{CB0000} \textbf{0.9844}} & 0.875                                  & 0.875                                  & 0.8906                                 & {\color[HTML]{CB0000} \textbf{0.8984}} \\ \hline
\end{tabular}
\label{Table:ablation_acc}
\end{table}

\begin{table}

\scriptsize
\centering
\caption{different metrics comparison with other different module settings for dataset2}
\begin{tabular}{cllllllll}
\hline
\textbf{light condition}                                                    & \multicolumn{4}{c}{\textbf{day}}                                                                                                                                  & \multicolumn{4}{c}{\textbf{night}}                                                                                                                                \\ \hline
\textbf{metrics}                                                            & \multicolumn{1}{c}{\textbf{precision}} & \multicolumn{1}{c}{\textbf{recall}}    & \multicolumn{1}{c}{\textbf{f1}}        & \multicolumn{1}{c}{\textbf{acc}}       & \multicolumn{1}{c}{\textbf{precision}} & \multicolumn{1}{c}{\textbf{recall}}    & \multicolumn{1}{c}{\textbf{f1}}        & \multicolumn{1}{c}{\textbf{acc}}       \\ \hline
\begin{tabular}[c]{@{}c@{}}no lighting \\ perception loss\end{tabular}      & 0.95801                                & 0.9629                                 & 0.9601                                 & 0.9757                                 & 0.9160                                 & 0.9421                                 & 0.9263                                 & 0.9531                                 \\
\begin{tabular}[c]{@{}c@{}}no lighting \\ condition perception\end{tabular} & {\color[HTML]{CB0000} \textbf{0.9609}} & {\color[HTML]{CB0000} \textbf{0.9864}} & {\color[HTML]{CB0000} \textbf{0.9725}} & 0.9740                                 & 0.9264                                 & 0.9716                                 & 0.9450                                 & 0.9656                                 \\
IMF                                                                        & 0.9580                                 & 0.9622                                 & 0.9601                                 & {\color[HTML]{CB0000} \textbf{0.9774}} & {\color[HTML]{CB0000} \textbf{0.9607}} & {\color[HTML]{CB0000} \textbf{0.9811}} & {\color[HTML]{CB0000} \textbf{0.9697}} & {\color[HTML]{CB0000} \textbf{0.9781}} \\ \hline
\end{tabular}
\label{Table:ablation_pvdf}

\end{table}

\subsubsection{Compared with different number of fusion layers}

In order to achieve the best fusion performance, we  further discuss the influence of different numbers of illumination-aware multi-modal fusion layers. Each fusion layer contains a residual block for both modalities respectively and a multi-modal fusion module. The terrains recognition results on dataset1 and dataset2 are demonstrated in the Table~\ref{Table:layernum} and Table~\ref{Table:layernum2},with the highest results at each condition is bold red.

In Table~\ref{Table:layernum} for dataset1, we observe that increasing the number of fusion layers leads to a gradual improvement in recognition accuracy across different working conditions. With only one fusion layer, the highest accuracy is achieved in just two conditions, whereas with two layers, the model achieves the highest recognition accuracy in five conditions. This indicates that increasing the number of fusion layers helps to more comprehensively extract complementary features between different modalities, thereby optimizing recognition performance. Considering both recognition performance and model complexity, we selected two fusion layers as the final model structure.

From Table~\ref{Table:layernum2}, we observe that when the number of layers is set to two, the recognition performance during daytime is slightly lower than that of other settings. However, the configuration with two layers achieves the best recognition results across all four metrics at night. Although using three layers yields the highest recall, F1-score, and accuracy under daytime conditions, its performance at night is significantly lower compared to those using two layers. Considering both daytime and nighttime recognition performance, we selected two fusion layers as the final model structure.

From different layer number settings we conclude that increasing the number of fusion layers significantly boosts the model's performance in road recognition tasks. The results suggest that deeper fusion enables the model to capture more complex and richer information, leading to higher accuracy across different datasets and conditions.

\begin{table}
\vspace{-0.3em}
\centering
\scriptsize

\caption{accuracy comparison with different numbers of fusion layers of dataset1}

\begin{tabular}{clllllllll}
\hline
\textbf{light condition} & \multicolumn{3}{c}{\textbf{noon}}                                                                                        & \multicolumn{3}{c}{\textbf{dusk}}                                                                                        & \multicolumn{3}{c}{\textbf{night}}                                                                                     \\ \hline
\textbf{speed(km/h)}     & \multicolumn{1}{c}{\textbf{10}}        & \multicolumn{1}{c}{\textbf{20}}        & \multicolumn{1}{c}{\textbf{30}}        & \multicolumn{1}{c}{\textbf{10}}        & \multicolumn{1}{c}{\textbf{20}}        & \multicolumn{1}{c}{\textbf{30}}        & \multicolumn{1}{c}{\textbf{10}}       & \multicolumn{1}{c}{\textbf{20}}        & \multicolumn{1}{c}{\textbf{30}}       \\ \hline
layer\_num=1             & 0.8438                                 & 0.8438                                 & 0.8906                                 & 0.875                                  & {\color[HTML]{000000} 0.9531}          & {\color[HTML]{CB0000} \textbf{0.8906}} & 0.7969                                & 0.9062                                 & {\color[HTML]{CB0000} \textbf{0.906}} \\
layer\_num=2             & {\color[HTML]{CB0000} \textbf{0.9219}} & 0.8438                                 & {\color[HTML]{CB0000} \textbf{0.9531}} & {\color[HTML]{CB0000} \textbf{0.9844}} & {\color[HTML]{CB0000} \textbf{0.9844}} & {\color[HTML]{000000} 0.875}           & {\color[HTML]{CB0000} \textbf{0.875}} & 0.8906                                 & 0.8984                                \\
layer\_num=3             & 0.8125                                 & 0.875                                  & 0.9062                                 & 0.9219                                 & 0.9062                                 & {\color[HTML]{CB0000} \textbf{0.8906}} & 0.7812                                & {\color[HTML]{CB0000} \textbf{0.9375}} & 0.8984                                \\
layer\_num=4             & 0.7969                                 & {\color[HTML]{CB0000} \textbf{0.9062}} & 0.9062                                 & 0.8438                                 & {\color[HTML]{000000} 0.9531}          & 0.875                                  & 0.8125                                & {\color[HTML]{CB0000} \textbf{0.9375}} & 0.8594                                \\ \hline
\end{tabular}
\label{Table:layernum}

\end{table}

\begin{table}
\vspace{-0.3em}
\centering
\scriptsize

\caption{differernt metrics comparison with different numbers of fusion layers of dataset2}

\begin{tabular}{ccccccccc}
\hline
\textbf{light condition}    & \multicolumn{4}{c}{\textbf{day}}                                                                                                                                  & \multicolumn{4}{c}{\textbf{night}}                                                                                                                                \\ \hline
\textbf{metrics} & \textbf{precision}                     & \textbf{recall}                        & \textbf{f1}                            & \textbf{acc}                           & \textbf{precision}                     & \textbf{recall}                        & \textbf{f1}                            & \textbf{acc}                           \\ \hline
layer\_num=1                   & 0.9637                                 & 0.9657                                 & 0.9644                                 & 0.9792                                 & 0.9209                                 & 0.9504                                 & 0.9330                                 & 0.9563                                 \\
lay\_num=2                     & 0.9580                                 & 0.9622                                 & 0.9601                                 & 0.9774                                 & {\color[HTML]{CB0000} \textbf{0.9607}} & {\color[HTML]{CB0000} \textbf{0.9811}} & {\color[HTML]{CB0000} \textbf{0.9697}} & {\color[HTML]{CB0000} \textbf{0.9781}} \\
layer\_num=3                   & 0.9648                                 & {\color[HTML]{CB0000} \textbf{0.9679}} & {\color[HTML]{CB0000} \textbf{0.9653}} & {\color[HTML]{CB0000} \textbf{0.9809}} & 0.9096                                 & 0.9409                                 & 0.9219                                 & 0.9500                                 \\
layer\_num=4                   & {\color[HTML]{CB0000} \textbf{0.9701}} & 0.9556                                 & 0.9616                                 & 0.9792                                 & 0.9270                                 & 0.9382                                 & 0.9299                                 & 0.9469                                 \\ \hline
\end{tabular}
\label{Table:layernum2}

\end{table}

\subsubsection{Compared with different hyperparameters}

We  also investigate the influence of $\lambda$ on the recognition accuracy of road terrains. We select $\lambda$ as 0, 0.2, 0.4, 0.6, 0.8 and 1.0 and the recognition results of both dataset1 and dataset2 are presented as Table.~\ref{Table:lambda} and   Table.~\ref{Table:lambda2},respectively, with the highest accuracy at each condition highlighted in bold red.

In Table.~\ref{Table:lambda} for dataset1, as the hyperparameter $\lambda$ increases, there is a notable improvement in the accuracy under various speed and lighting conditions. When $\lambda=0$, the highest accuracy is achieved in only two conditions. However, when $\lambda=1.0$ , the highest accuracy is achieved in four and five conditions, respectively. This indicates that increasing $\lambda$ can enhance the weight of the illumination perception loss, thereby improving road terrain recognition results.

In Table.~\ref{Table:lambda2} for dataset2, a similar trend is observed, with the model's precision, recall, F1-score, and accuracy improving as $\lambda$ increases. For instance, at $\lambda$ = 1.0, the algorithm achieves the highest accuracy at  both day and night, along with the highest precision (0.9781), recall (0.9607) and f1-score(0.9697) under nighttime condition. This suggests that $\lambda$ plays a critical role in balancing the algorithm's performance, particularly in terms of its ability to generalize across different lighting scenarios.

In conclusion, $\lambda$ is able to adjust the weight of illumination loss and proper value of $\lambda$ can achieve a balance between illumiantion perception and terrain classification. Finally, we select $\lambda$ = 1.0 to train our algorithm.

\begin{table*}[]
\vspace{-0.3em}
\centering
\scriptsize

\caption{accuracy comparison with different hyperparameter values of dataset1}

\begin{tabular}{cccccccccc}
\hline
\textbf{light condition} & \multicolumn{3}{c}{\textbf{noon}}                                                                                       & \multicolumn{3}{c}{\textbf{dusk}}                                                                                        & \multicolumn{3}{c}{\textbf{night}}                                                                                       \\ \hline
\textbf{speed(km/h)}     & \textbf{10}                            & \textbf{20}                           & \textbf{30}                            & \textbf{10}                            & \textbf{20}                            & \textbf{30}                            & \textbf{10}                            & \textbf{20}                            & \textbf{30}                            \\ \hline
$\lambda$=0.0               & 0.8906                                 & 0.7812                                & 0.875                                  & 0.8906                                 & {\color[HTML]{000000} 0.9219}          & {\color[HTML]{000000} 0.9062}          & {\color[HTML]{CB0000} \textbf{0.9062}} & {\color[HTML]{CB0000} \textbf{0.9219}} & {\color[HTML]{000000} 0.875}           \\
$\lambda$=0.2               & {\color[HTML]{000000} 0.9062}          & 0.8594                                & {\color[HTML]{000000} 0.9219}          & {\color[HTML]{000000} 0.9062}          & 0.875                                  & {\color[HTML]{000000} 0.8906}          & {\color[HTML]{000000} 0.8438}          & 0.9375                                 & 0.8906                                 \\
$\lambda$=0.4               & 0.8594                                 & 0.7812                                & 0.8906                                 & 0.8594                                 & 0.9062                                 & {\color[HTML]{000000} 0.8594}          & {\color[HTML]{CB0000} \textbf{0.9062}} & {\color[HTML]{CB0000} \textbf{0.9219}} & 0.8672                                 \\
$\lambda$=0.6               & 0.875                                  & {\color[HTML]{CB0000} \textbf{0.875}} & 0.8906                                 & 0.9375                                 & 0.9375                                 & 0.8594                                 & 0.8125                                 & 0.9062                                 & 0.8672                                 \\
$\lambda$=0.8               & {\color[HTML]{CB0000} \textbf{0.9219}} & 0.7812                                & 0.8906                                 & 0.9219                                 & 0.9219                                 & {\color[HTML]{CB0000} \textbf{0.9375}} & 0.8281                                 & 0.9062                                 & 0.875                                  \\
$\lambda$=1.0               & {\color[HTML]{CB0000} \textbf{0.9219}} & {\color[HTML]{000000} 0.8438}         & {\color[HTML]{CB0000} \textbf{0.9531}} & {\color[HTML]{CB0000} \textbf{0.9844}} & {\color[HTML]{CB0000} \textbf{0.9844}} & 0.875                                  & 0.875                                  & {\color[HTML]{000000} 0.8906}          & {\color[HTML]{CB0000} \textbf{0.8984}} \\ \hline
\end{tabular}
\label{Table:lambda}
\end{table*}

\begin{table}
\vspace{-0.3em}
\scriptsize

\centering

\caption{differernt metrics comparison with different hyperparameter values of dataset2}

\begin{tabular}{ccccccccc}
\hline
\textbf{lighting condition} & \multicolumn{4}{c}{\textbf{day}}                                                                                                                                  & \multicolumn{4}{c}{\textbf{light}}                                                                                                                                \\ \hline
\textbf{metrics}            & \textbf{precision}                     & \textbf{recall}                        & \textbf{f1}                            & \textbf{acc}                           & \textbf{precision}                     & \textbf{recall}                        & \textbf{f1}                            & \textbf{acc}                           \\ \hline
$\lambda$=0.0                  & 0.9580                                 & {\color[HTML]{CB0000} \textbf{0.9629}} & {\color[HTML]{CB0000} \textbf{0.9601}} & 0.9757                                 & 0.9160                                 & 0.9421                                 & 0.9263                                 & 0.9531                                 \\
$\lambda$=0.2                  & 0.9557                                 & 0.9623                                 & 0.9583                                 & {\color[HTML]{CB0000} \textbf{0.9774}} & 0.9065                                 & 0.9399                                 & 0.9204                                 & 0.9469                                 \\
$\lambda$=0.4                  & {\color[HTML]{333333} 0.9510}          & {\color[HTML]{333333} 0.9513}          & {\color[HTML]{333333} 0.9510}          & {\color[HTML]{333333} 0.974}           & 0.9058                                 & 0.9256                                 & 0.9120                                 & 0.9469                                 \\
$\lambda$=0.6                  & {\color[HTML]{CB0000} \textbf{0.9585}} & 0.9576                                 & {\color[HTML]{333333} 0.9579}          & {\color[HTML]{333333} 0.9774}          & 0.9243                                 & 0.931                                  & 0.9236                                 & 0.9438                                 \\
$\lambda$=0.8                  & 0.9438                                 & 0.9558                                 & 0.9485                                 & 0.9722                                 & 0.9119                                 & 0.9440                                 & 0.9227                                 & 0.9469                                 \\ \hline
$\lambda$=1.0                  & {\color[HTML]{333333} 0.9580}          & 0.9622                                 & {\color[HTML]{CB0000} \textbf{0.9601}} & {\color[HTML]{CB0000} \textbf{0.9774}} & {\color[HTML]{CB0000} \textbf{0.9607}} & {\color[HTML]{CB0000} \textbf{0.9812}} & {\color[HTML]{CB0000} \textbf{0.9697}} & {\color[HTML]{CB0000} \textbf{0.9781}} \\ \hline
\end{tabular}
\label{Table:lambda2}

\end{table}

\subsubsection{Time and Computational Resource Consumption of Different methods}
 We further added comparisons to illustrate the differences in computational efficiency among the various methods. The performance of various methods on Dataset1 is presented in Table \ref{Table:time_dataset1}. CEN exhibits the highest inference time (0.5369 s), with a substantial parameter count (98.6283M) and FLOPs (118.2047G), reflecting its inefficiency. In contrast, our proposed IMF achieves a significantly lower inference time of 0.1853 s, outperforming Transformer-based methods like DSF and TKF, and closely matching efficient CNN-based methods such as MMTM and EIP. While DSF has the fewest parameters (0.1858M), its FLOPs are the highest (397.0832G), whereas IMF maintains a balanced 2.8929M parameters and 11.6636G FLOPs, far more efficient than other baselines. CPU usage across methods is similar, ranging from 16.4434MB (DSF) to 17.4407MB (TKF), with IMF at 16.9497MB.

On Dataset2, as shown in the Table \ref{Table:time_dataset1}, CEN again underperforms with an inference time of 2.5144 s, 100.7478M parameters, and 118.216G FLOPs. IMF, however, achieves an impressive 0.1565 s inference time, surpassing even the fastest method. With 2.899M parameters and 11.664G FLOPs, our model remains far more efficient than other baselines. Although CPU usage varies widely (e.g., DSF at 115.1234MB), our method’s 76.4494MB is comparable to most methods. Overall, IMF consistently demonstrates superior efficiency and speed across both datasets, making it a highly competitive choice for resource-constrained applications.

\begin{table}
\vspace{-0.3em}
\centering
\scriptsize
\caption{Time and computational resource consumption of Different methods on dataset1.}
\begin{tabular}{cccccc}
\hline
\multicolumn{2}{c}{\textbf{time and computing}}                              & \textbf{time(s)} & \textbf{paramemers(M)} & \textbf{flops} & \textbf{cpu(MB)} \\ \hline
\multirow{3}{*}{\textbf{channel-exchanging   based on CNN}}         & MMTM   & 0.1824        & 2.9223                       & 12.3584        & 16.9772          \\
& CEN    & 0.5369        & 98.6283                      & 118.2047       & 16.773           \\
& EIP    & 0.1864        & 5.3448                       & 9.1346         & 16.5952          \\ \cline{1-1}
\multirow{5}{*}{\textbf{channel-exchanging   based on Transformer}} & mbt    & 0.1871        & 0.9296                       & 10.1827        & 17.1257          \\
& MFT    & 0.185         & 1.844                        & 25.7407        & 16.6471          \\
& TKF    & 0.3116        & 16.2675                      & 172.7836       & 17.4407          \\
& DSF    & 0.507         & 0.1858                       & 397.0832       & 16.4434          \\
& mmsf   & 0.2413        & 6.8446                       & 79.0884        & 16.8762          \\ \cline{1-1}
\multirow{3}{*}{\textbf{aggragation-based   method}}                & early  & 0.1916        & 2.7571                       & 31.2922        & 16.7084          \\
& middle & 0.192         & 2.6628                       & 24.4486        & 16.687           \\
 & late   & 0.1921        & 3.4207                       & 31.3347 & 16.7163          \\ \hline
\multicolumn{2}{c}{\textbf{IMF}}                                             & 0.1853        & 2.8929                       & 11.6636        & 16.9497          \\ \hline
\end{tabular}
\label{Table:time_dataset1}
\end{table}

\begin{table}[h]
\vspace{-0.3em}
\centering
\scriptsize
\caption{Time and computational resource consumption of Different methods on dataset1.}
\label{tab:dataset2}
\begin{tabular}{cccccc}
\hline
\multicolumn{2}{c}{\textbf{time and computing}}                              & \multicolumn{1}{c}{\textbf{time}} & \multicolumn{1}{c}{\textbf{paramemers(M)}} & \multicolumn{1}{c}{\textbf{flops}} & \multicolumn{1}{c}{\textbf{cpu(MB)}} \\ \hline
\multirow{3}{*}{\textbf{channel-exchanging   based on CNN}}         & MMTM   & 0.3122                            & 2.9283                                           & 12.3588                            & 75.6314                              \\
& CEN    & 2.5144                            & 100.7478                                         & 118.2156                           & 75.732                               \\
& EIP    & 0.3351                            & 5.3478                                           & 9.1348                             & 76.2885                              \\ \cline{1-1}
\multirow{5}{*}{\textbf{channel-exchanging   based on Transformer}} & mbt    & 0.3769                            & 0.9307                                           & 10.1827                            & 72.3541                              \\
& MFT    & 0.3436                            & 1.8442                                           & 25.7407                            & 76.9228                              \\
& TKF    & 0.9748                            & 17.4525                                          & 172.872                            & 79.4666                              \\
& DSF    & 1.6192                            & 0.1865                                           & 397.0832                           & 115.1234                             \\
 & mmsf   & 0.6865                            & 6.8454                                           & 76.3134                            & 79.0884                              \\ \cline{1-1}
\multirow{3}{*}{\textbf{aggragation-based   method}}                & early  & 0.3736                            & 2.7606                                           & 31.2925                            & 76.2563                              \\
 & middle & 0.3545                            & 2.6668                                           & 24.4488                            & 75.93                                \\
& late   & 0.3807                            & 3.4241                                           & 31.3349                            & 72.718                               \\ \hline
\multicolumn{2}{c}{\textbf{IMF}}                                             & 0.1565                            & 2.899                                            & 11.664                             & 76.4494                              \\ \hline
\end{tabular}
\label{Table:time_dataset2}
\end{table}

\subsubsection{Visualization for wrong predicted data}

Furthermore, for both datasets, we extract the misclassified original data for visualization and qualitative analysis.

\begin{figure}[htp]
    \centering
    \includegraphics[width=12cm]{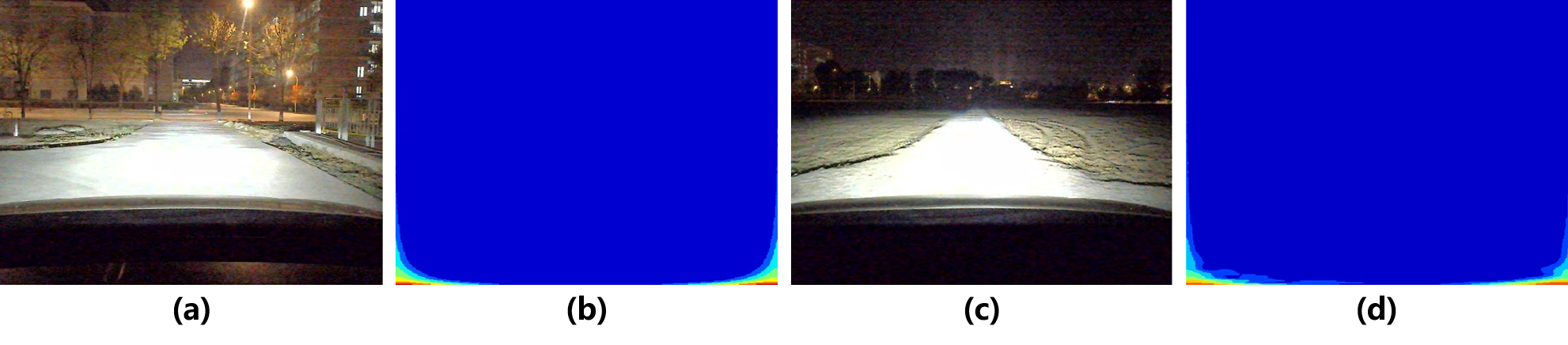}
    \caption{(a)-(b):cement road at night; (c)-(d): gravel road at night.}
    \label{fig:wrong_prd_acc}
\end{figure}

As shown in Fig \ref{fig:wrong_prd_acc}, the images depict some misclassified samples from Dataset1. Specifically, Fig (a) and (b) correspond to cement roads but were misclassified as asphalt roads. This mis-classification may be attributed to nighttime conditions, where the vehicle's headlights illuminate the road surface, causing its features to appear blurred in the images. Additionally, the corresponding proprioceptive data, i.e., the spectrograms of acceleration data, also exhibit relatively smooth patterns. Since the features in both modalities appear indistinct, the classification result was incorrect. Fig (c) and (d) correspond to gravel roads but were misclassified as cement roads. Similarly, the vehicle's headlights caused overexposure in the images, resulting in the loss of road surface feature information. The spectrograms of the corresponding proprioceptive data exhibit slight fluctuations at the edges but remain relatively minor, failing to provide effective feature inputs. Consequently, this led to misclassification.

\begin{figure}[htp]
    \centering
    \includegraphics[width=12cm]{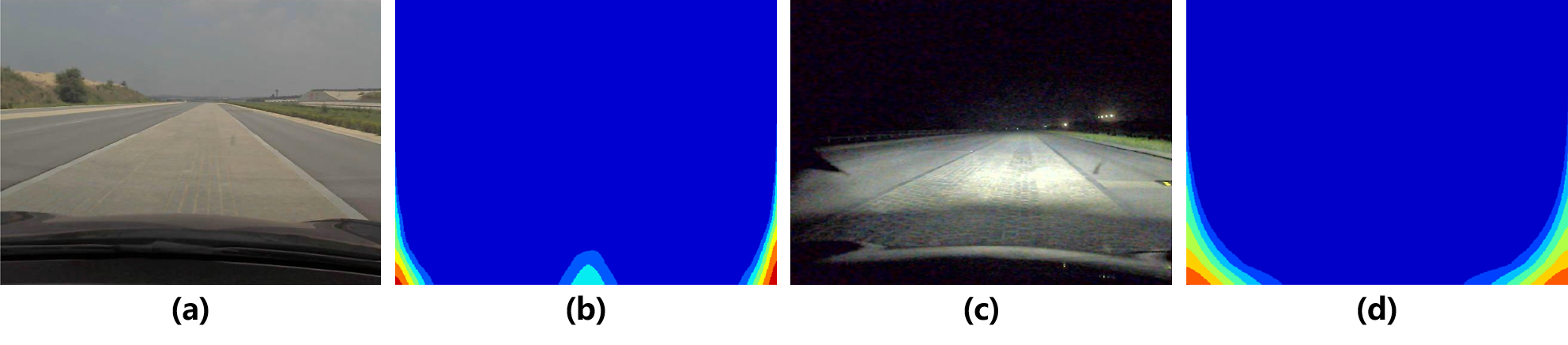}
    \caption{(a)-(b):brick road at daytime; (c)-(d): brick road at night.}
    \label{fig:wrong_prd_pvdf}
\end{figure}

As shown in Fig \ref{fig:wrong_prd_pvdf}, the images depict some misclassified samples from dataset2. Specifically, Fig (a) and (b) correspond to brick roads but were misclassified as cement roads. This mis-classification may be due to a prominent peak in the spectrogram of the proprioceptive data, leading the algorithm to incorrectly estimate the road type. Fig (c) and (d) also correspond to brick roads but were misclassified as patched asphalt. On one hand, nighttime driving caused overexposure due to the vehicle's headlights illuminating the road, resulting in the loss of most visual information. On the other hand, the spectrogram closely resembles the characteristics of patched asphalt, causing the algorithm to misjudge the classification.

Overall, the misclassifications are primarily caused by overexposure in images due to nighttime driving, which results in the loss of most road surface features. As shown in Table \ref{Table:baselinecompare} and Table \ref{Table:baselinepvdf}, the recognition accuracy under different conditions still outperforms other methods, and the mis-classification probability remains within an acceptable range. In future work, we will explore improved image acquisition methods to reduce overexposure in nighttime road images and enhance the quality of the original data.

\section{Conclusion}
\label{Sec:conclusion}

In this study, we propose an illumination-aware multi-modal fusion network (IMF) to improve the real-time perception of road terrains for autonomous vehicles (AVs) under varying lighting conditions. By integrating exteroceptive and proprioceptive sensing and dynamically adjusting their fusion weights based on estimated illumination features, IMF effectively mitigates the limitations of conventional visual-based methods, which are highly susceptible to illumination and weather variations. Additionally, the pre-training strategy and loss of the illumination perception sub-network contribute to more effective learning and optimization. Experimental results confirm that IMF outperforms state-of-the-art methods and highlights the benefits of multi-modal fusion over single-modality approaches.

Our work demonstrates the effectiveness of illumination perception in multi-modal fusion for real-world autonomous driving scenarios. The proposed illumination-aware fusion strategy can be extended to other tasks, such as object detection under adverse lighting. However, our work still has limitations, including potential performance degradation under extreme weathers and the lack of consideration of other critical road surface characteristics, such as friction coefficient and anomalies. Future work should incorporate these additional features to enhance robustness and improve generalization to real-world driving scenarios. We believe IMF provides a solid foundation for further advancements in multi-modal perception for AVs.

\bibliographystyle{elsarticle-num}
\bibliography{arxiv}

\end{document}